\ifdefined\pdfminorversion
\fi
\documentclass[10pt]{article}

\usepackage[final,eandd,nonatbib]{neurips_2026}
\renewcommand{\acksection}{\section*{Acknowledgments}}
\usepackage[numbers,sort&compress]{natbib}

\usepackage[utf8]{inputenc}
\usepackage[T1]{fontenc}
\usepackage{hyperref}
\usepackage{url}
\usepackage{booktabs}
\usepackage{amsfonts}
\usepackage{amsmath}
\usepackage{amssymb}
\usepackage{nicefrac}
\usepackage{microtype}
\usepackage{xcolor}
\usepackage{graphicx}
\usepackage{multirow}
\usepackage{enumitem}
\usepackage{wrapfig}
\usepackage{subcaption}
\usepackage{float}
\usepackage{makecell}
\usepackage{listings}

\lstdefinestyle{promptstyle}{
  basicstyle=\ttfamily\footnotesize,
  breaklines=true,
  breakatwhitespace=false,
  columns=fullflexible,
  keepspaces=true,
  showstringspaces=false,
  upquote=true,
  literate={`}{{\textasciigrave}}1
}

\newcommand{\pointcloud}{\mathcal{P}}    % point cloud
\newcommand{\grasppose}{\mathbf{g}}      % grasp pose
\newcommand{\graspset}{\mathcal{C}}      % set of candidate grasps
\newcommand{\traj}{\boldsymbol{\tau}}    % trajectory

\newcommand{\argmax}{\operatornamewithlimits{argmax}}

\newcommand{\figTeaser}{%
\begin{figure}[h]
\centering
\includegraphics[width=\linewidth,trim=50 90 190 90,clip]{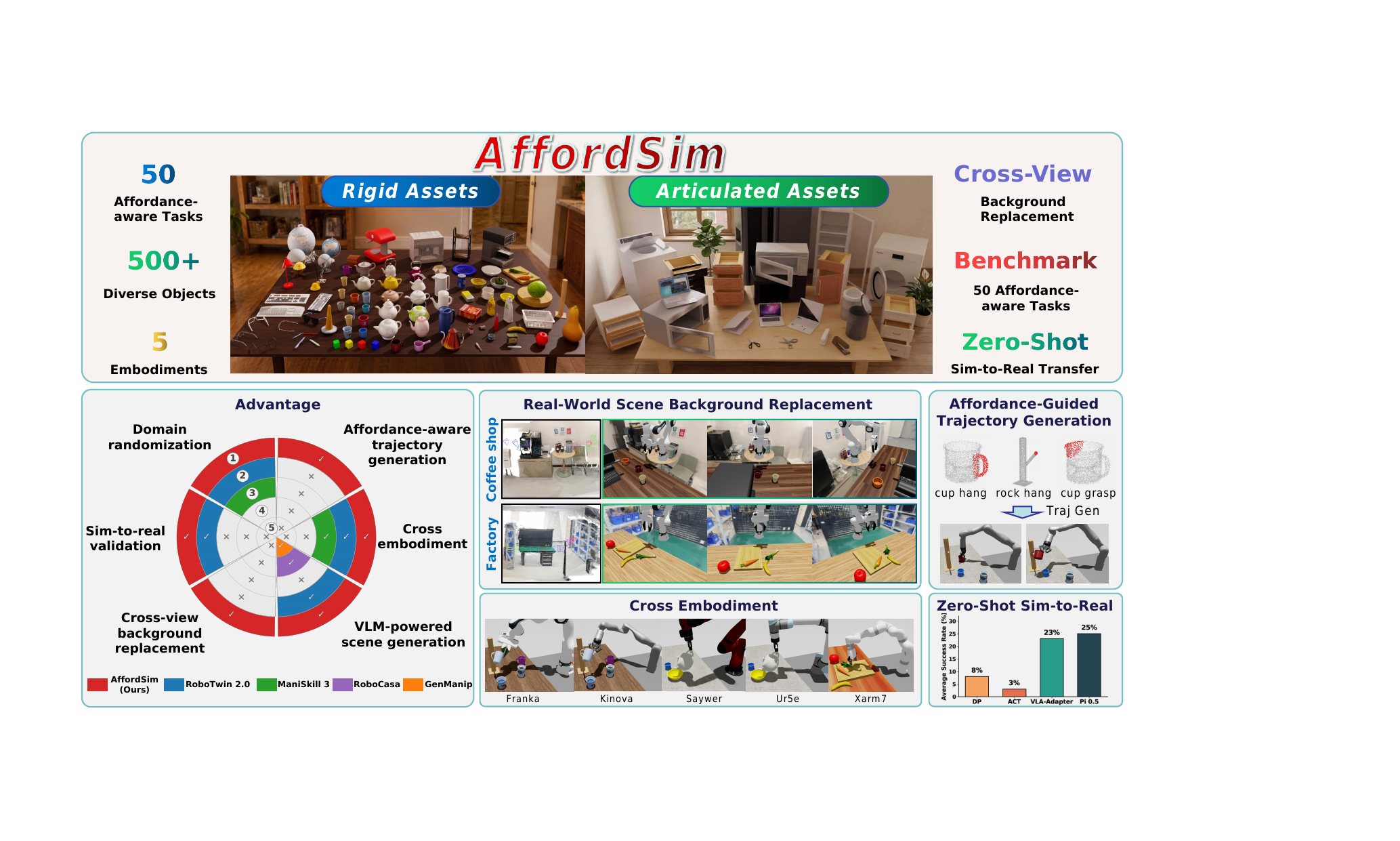}
\caption{\textbf{Overview of AffordSim.} A scalable affordance-aware manipulation benchmark of 50 tasks, 500+ rigid and articulated objects, and 5 robot embodiments, paired with cross-viewpoint background replacement and affordance-guided trajectory generation.}
\label{fig:teaser}
\end{figure}
}

\newcommand{\figOverview}{%
\begin{figure*}[t]
\centering
\includegraphics[width=0.98\textwidth,trim=92 645 207 196,clip]{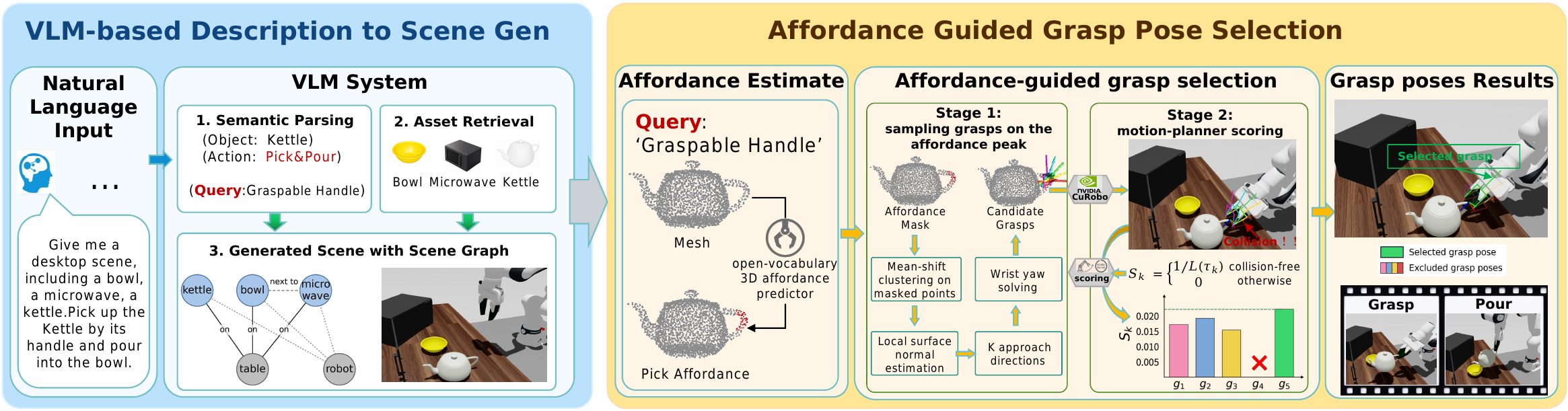}
\caption{\textbf{Affordance-guided data generation pipeline of AffordSim.} The full module breakdown and data flow are described in Section~\ref{sec:method}. In the Stage~2 bar chart, the x-axis lists the candidate grasps $g_k$ generated in Stage~1 and the y-axis gives their motion-planner score $S_k$, with the red cross marking the colliding candidate ($S_k = 0$) and the green dashed line indicating the maximum-score grasp ultimately selected.}
\label{fig:overview}
\vspace{-1em}
\end{figure*}
}

\newcommand{\figAffordanceWhy}{%
\begin{figure*}[!htbp]
\centering
\includegraphics[width=\textwidth,trim=0 110 0 95,clip]{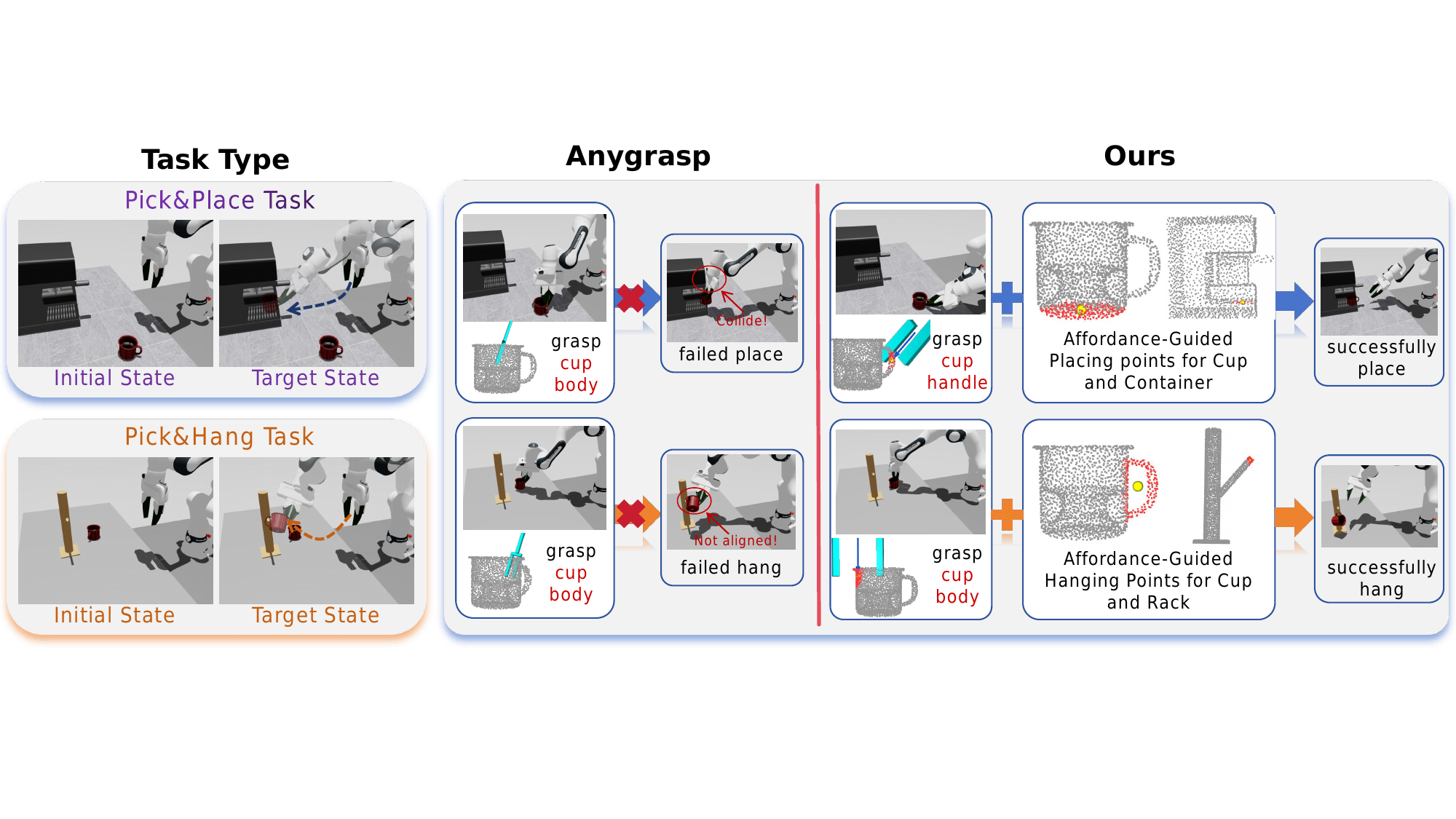}
\caption{\textbf{Why affordance is essential for trajectory generation.} Comparison of AnyGrasp and our affordance-guided grasping on two affordance-dependent tasks, pick \& place and hang.}
\label{fig:affordance_why}
\end{figure*}
}

\newcommand{\figDomainRand}{%
\begin{figure*}[t]
\centering
\includegraphics[width=\textwidth]{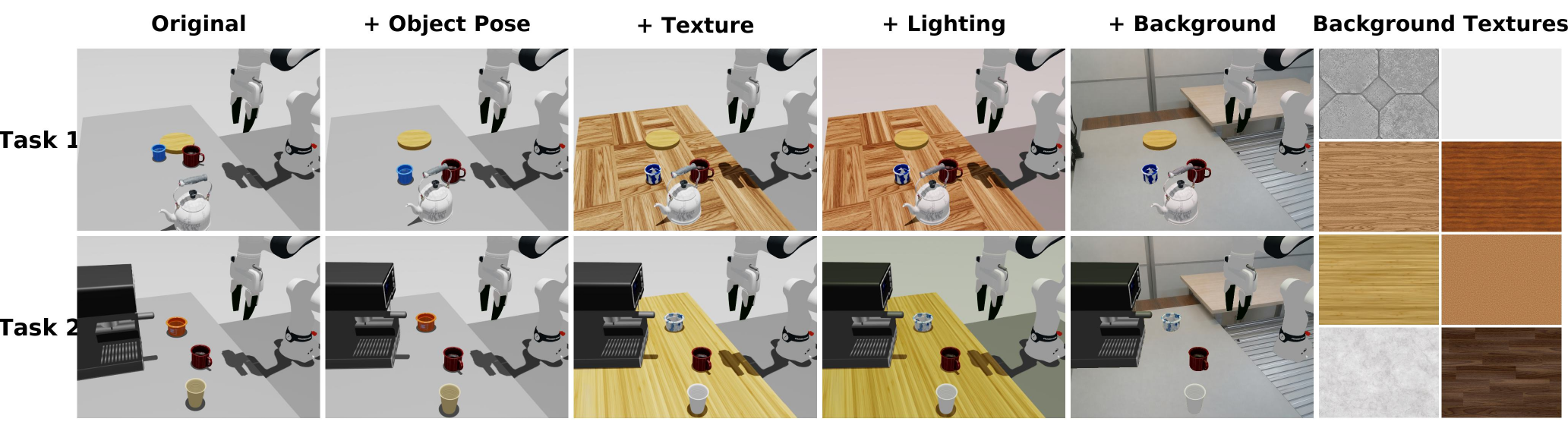}
\caption{\textbf{Domain randomization in AffordSim.} Each column shows the cumulative effect of adding one randomization axis: object pose, surface texture, lighting, and background. The rightmost column shows example background textures used for randomization. Two representative tasks are shown across rows.}
\label{fig:domain_rand}
\vspace{-1em}
\end{figure*}
}

\newcommand{\figTaskGallery}{%
\begin{figure*}[t]
\centering
\includegraphics[width=0.98\textwidth]{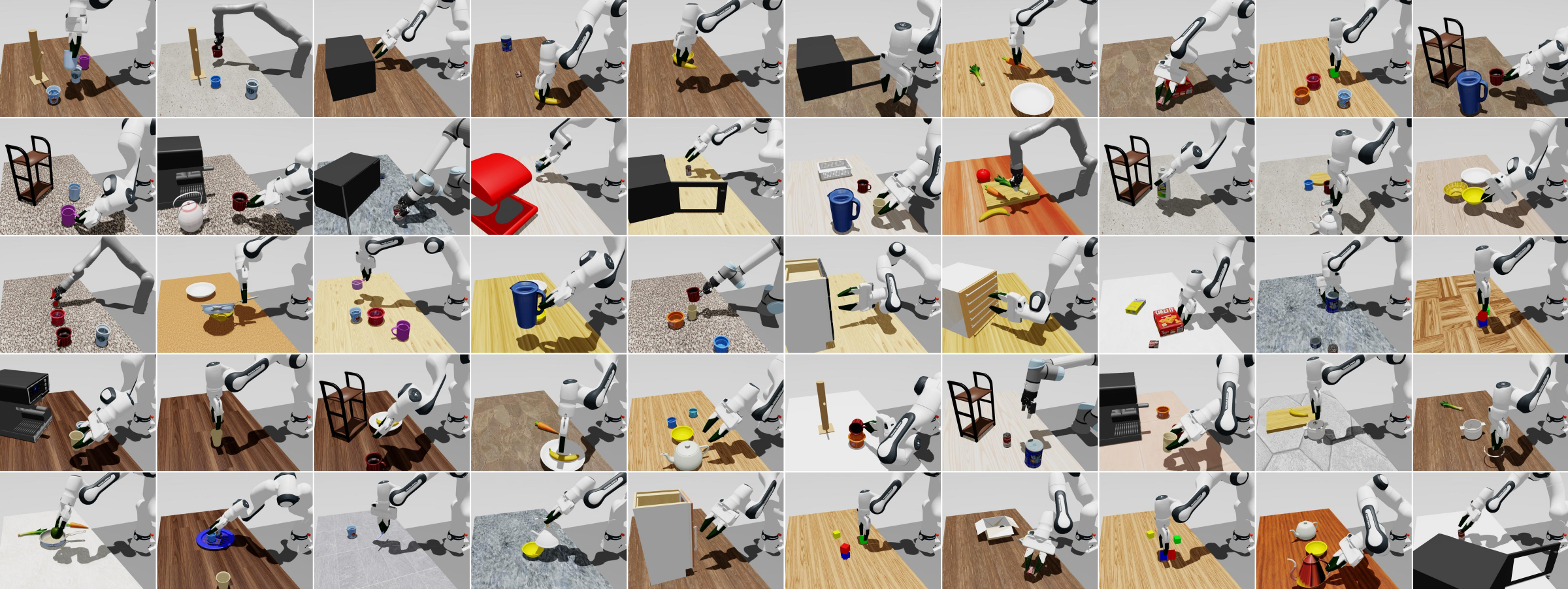}
\caption{\textbf{Task gallery.} Representative frames from all 50 tasks of the AffordSim benchmark, grouped into seven manipulation categories: pick \& place (20), open/close (2), pull/push (5), hang (2), pour (7), stack (4), and long-horizon composite (10).}
\label{fig:task_gallery}
\vspace{-1em}
\end{figure*}
}

\newcommand{\figCrossEmbodiment}{%
\begin{figure*}[!htbp]
\centering
\includegraphics[width=\textwidth]{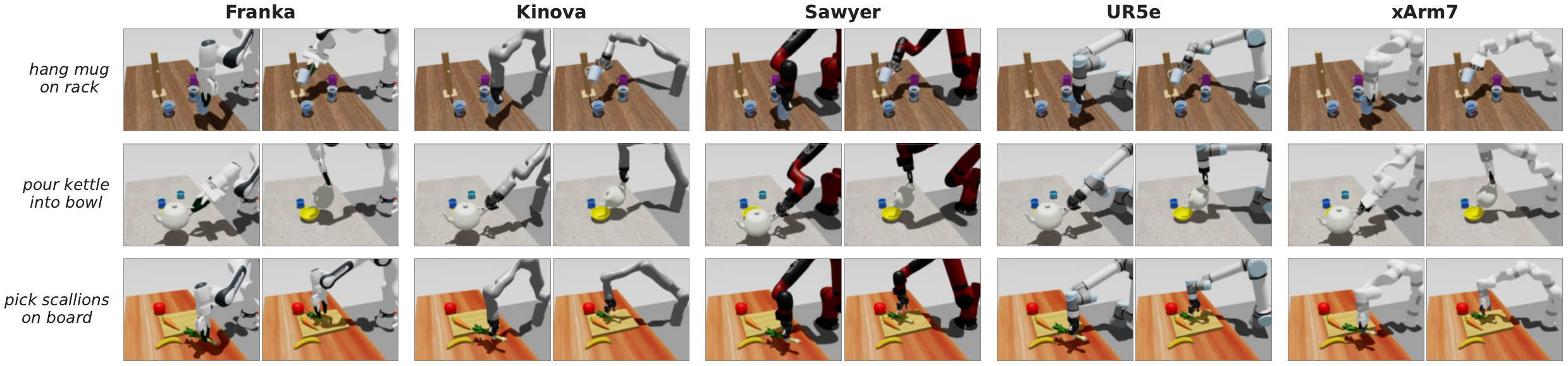}
\caption{\textbf{Cross-embodiment support.} Start and end frames of three representative affordance-dependent tasks (\textit{hang mug on rack}, \textit{pour kettle into bowl}, \textit{pick scallions on board}) generated by AffordSim for each of the five supported embodiments.}
\label{fig:cross_embodiment}
\end{figure*}
}

\newcommand{\figSimToReal}{%
\begin{figure}[H]
\centering
\setlength{\tabcolsep}{2pt}
\renewcommand{\arraystretch}{0.4}
\begin{tabular}{cccc}
\includegraphics[width=0.23\linewidth]{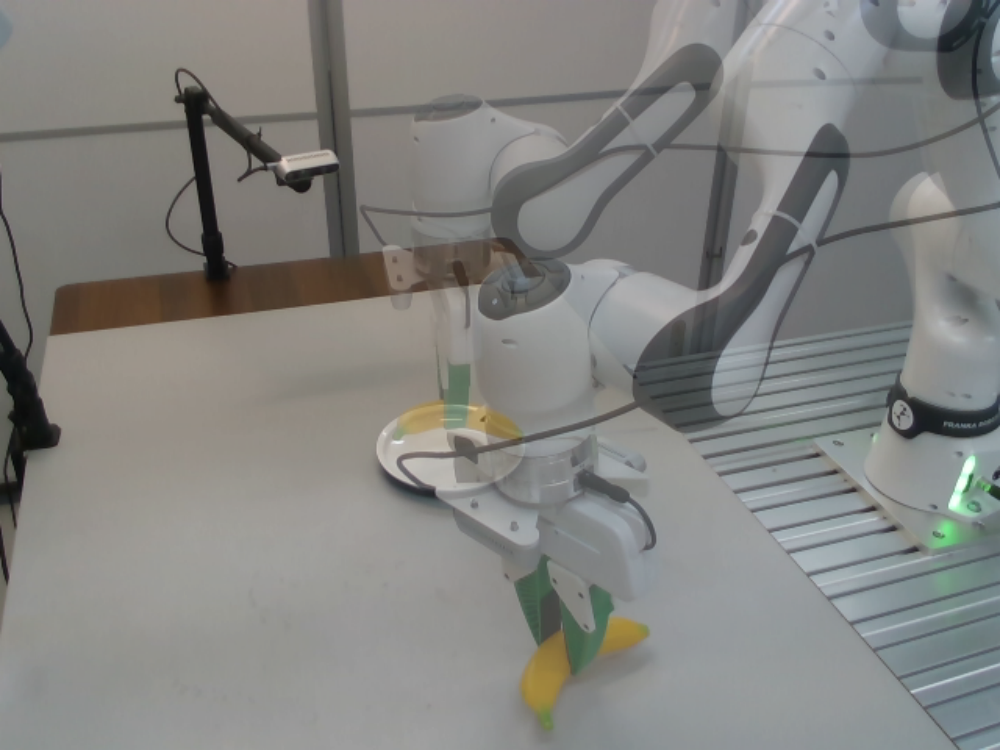} &
\includegraphics[width=0.23\linewidth]{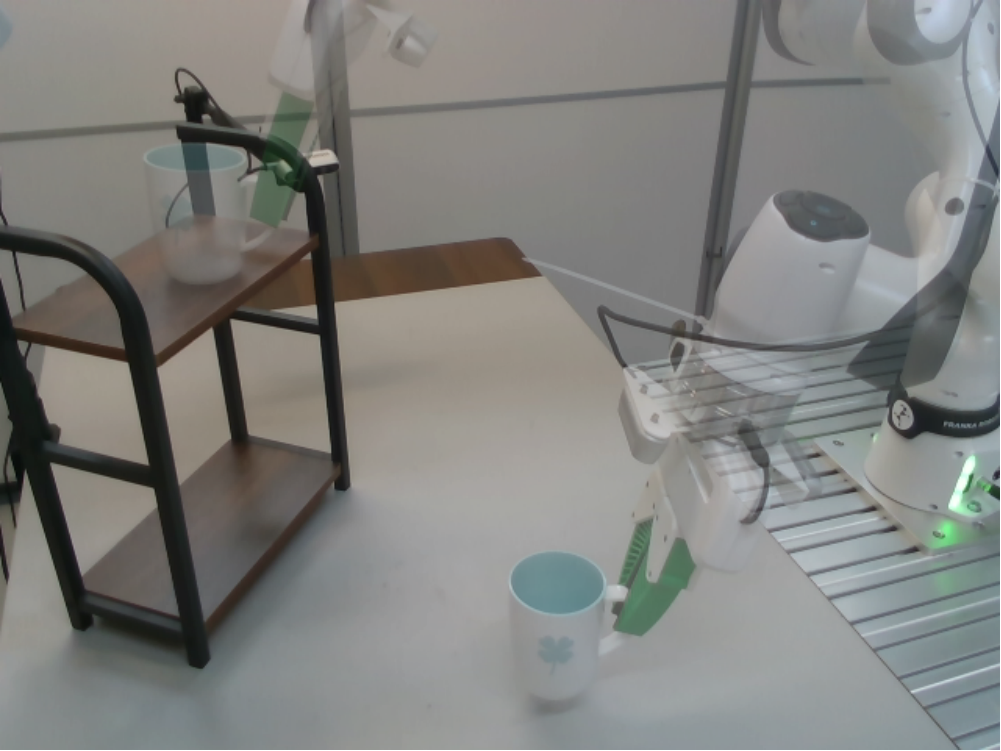} &
\includegraphics[width=0.23\linewidth]{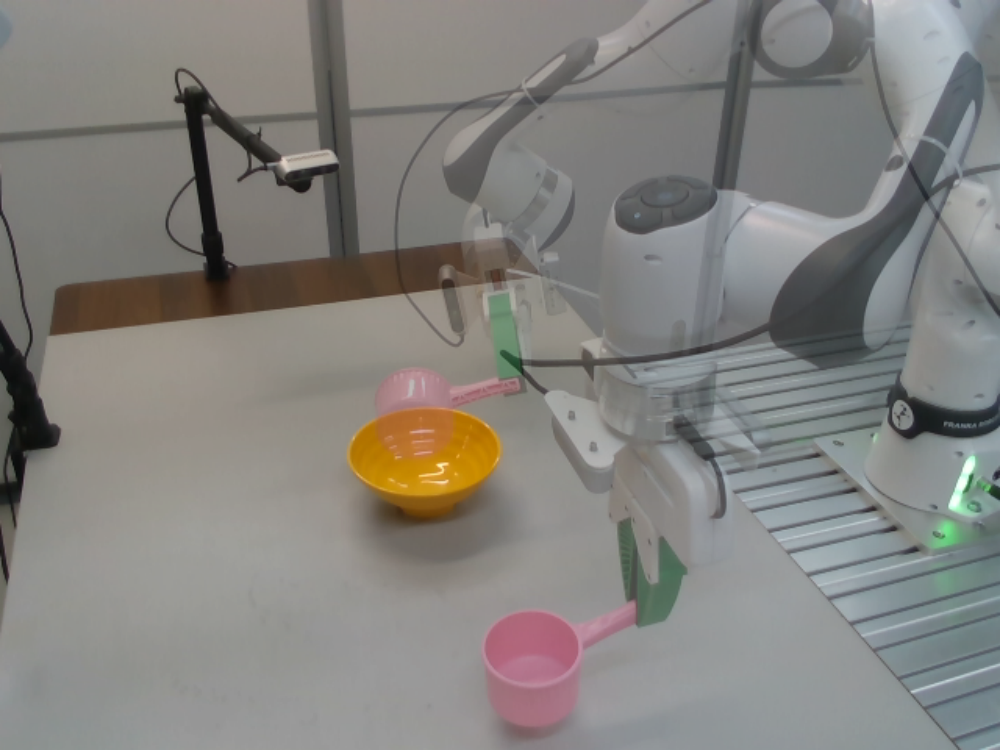} &
\includegraphics[width=0.23\linewidth]{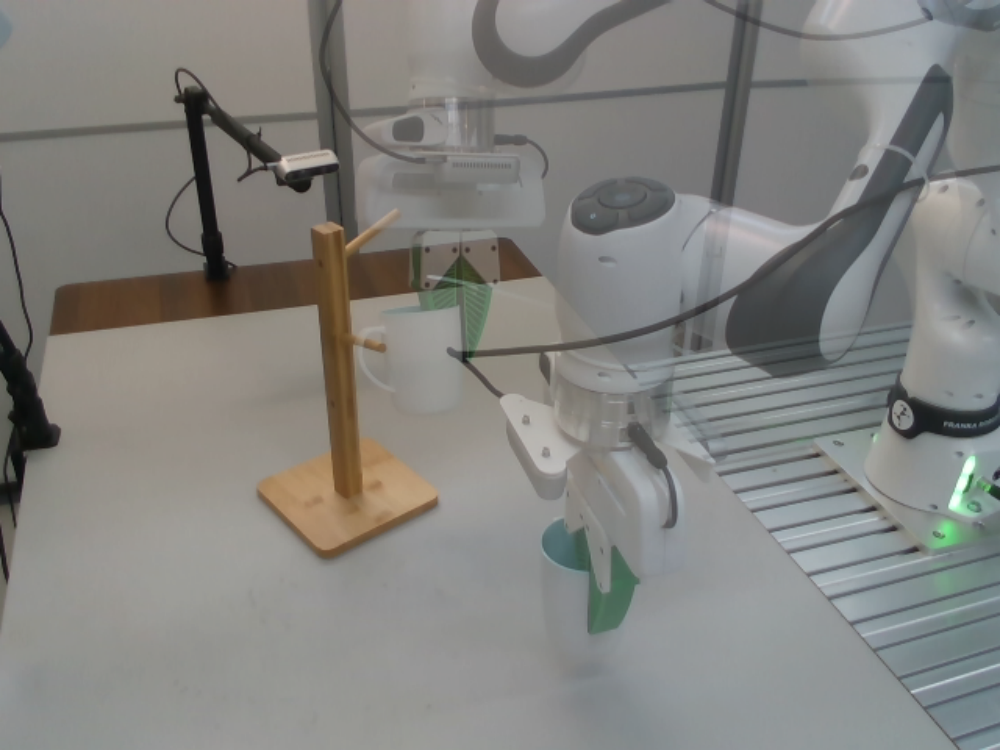} \\[1pt]
{\scriptsize\textit{pick banana place plate}} &
{\scriptsize\textit{pick cup place shelf}} &
{\scriptsize\textit{pour scoop into bowl}} &
{\scriptsize\textit{hang red mug on rack}} \\
\end{tabular}
\caption{\textbf{Zero-shot sim-to-real deployment.} The four real robot tasks in Table~\ref{tab:sim2real}, executed by Pi\,0.5 on a Franka FR3 with no real world fine tuning.}
\label{fig:sim2real}
\vspace{-1em}
\end{figure}
}

\newcommand{\figDAThreeInputs}{%
\begin{figure}[!htbp]
\centering
\setlength{\tabcolsep}{1.5pt}
\renewcommand{\arraystretch}{0.5}
\begin{tabular}{@{}cccccc@{}}
\rotatebox{90}{\,\,\scriptsize Coffee corner} &
\includegraphics[width=0.18\linewidth]{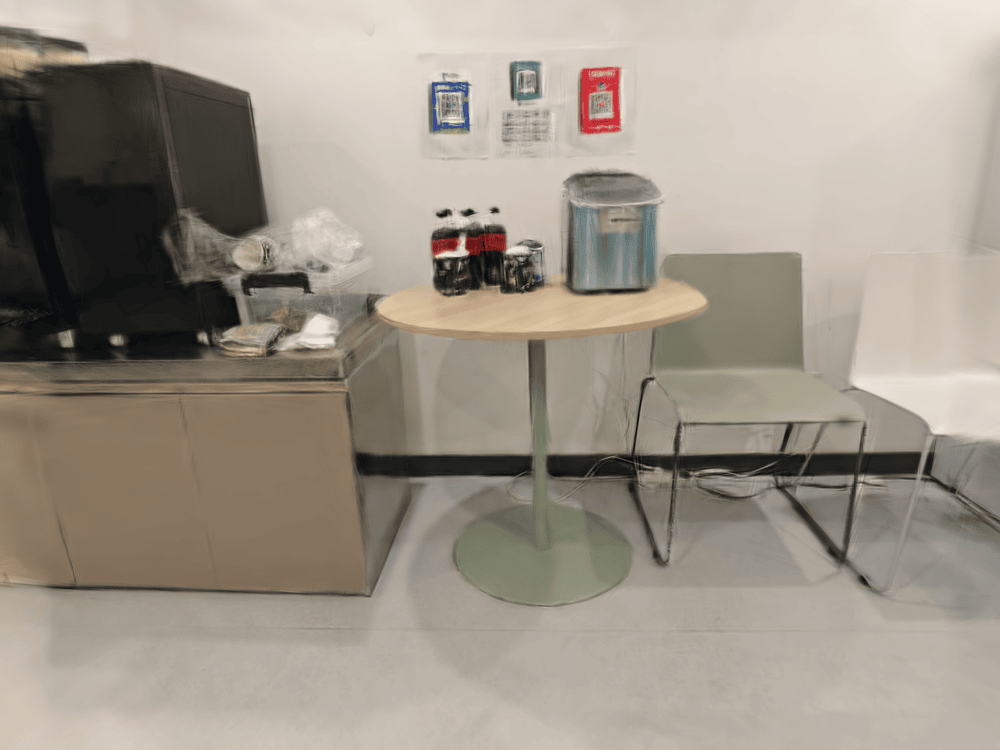} &
\includegraphics[width=0.18\linewidth]{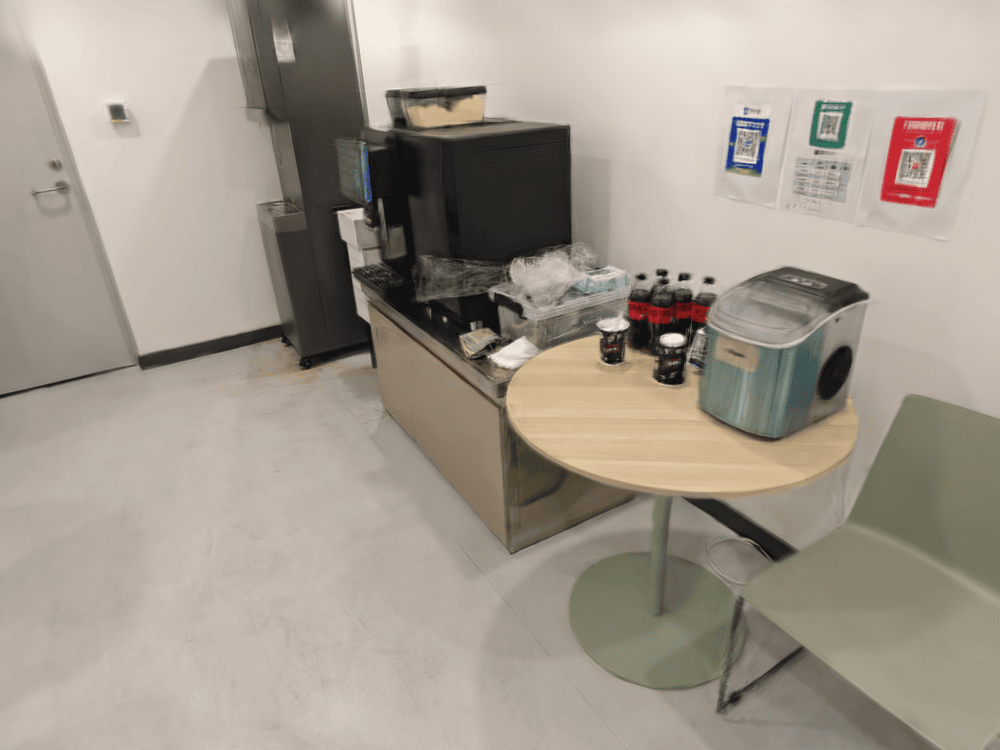} &
\includegraphics[width=0.18\linewidth]{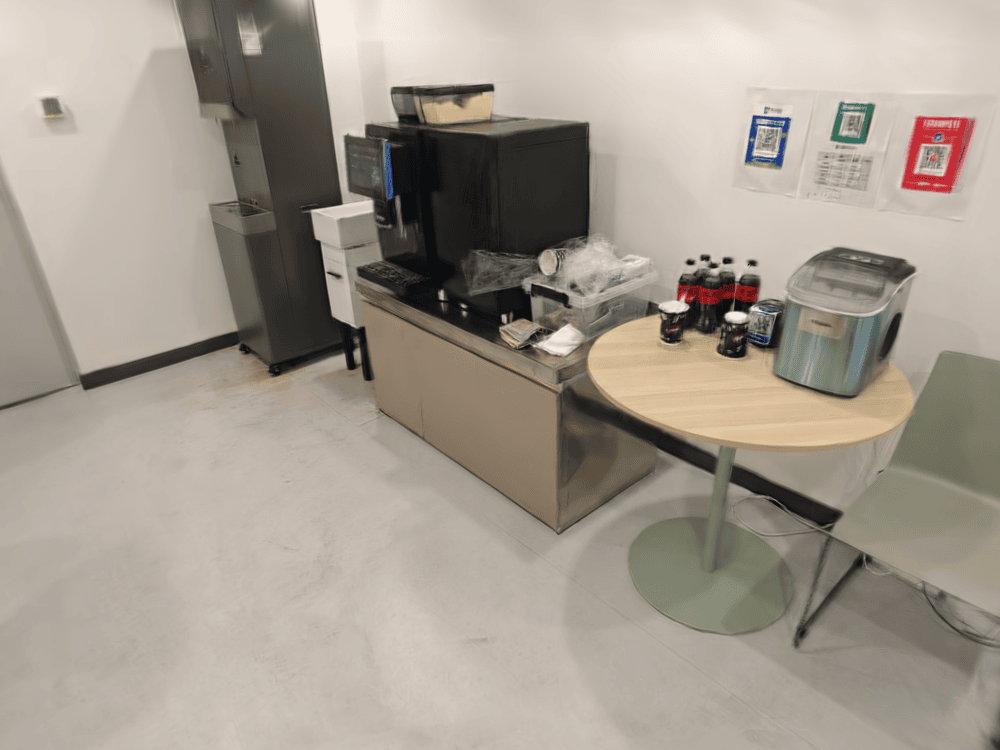} &
\includegraphics[width=0.18\linewidth]{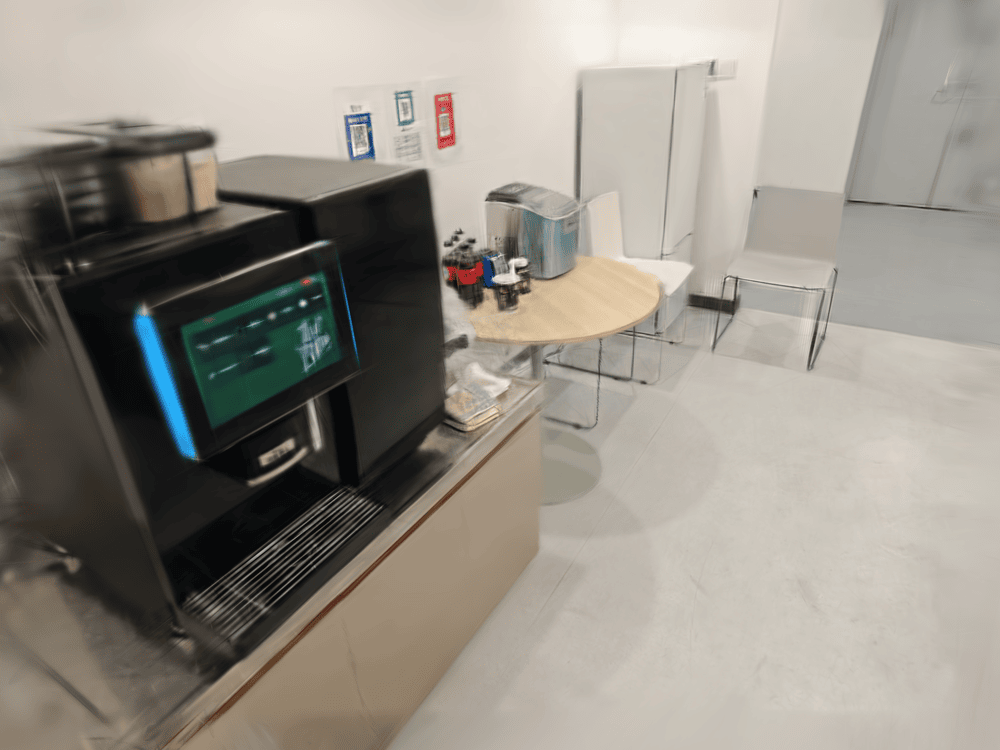} &
\includegraphics[width=0.18\linewidth]{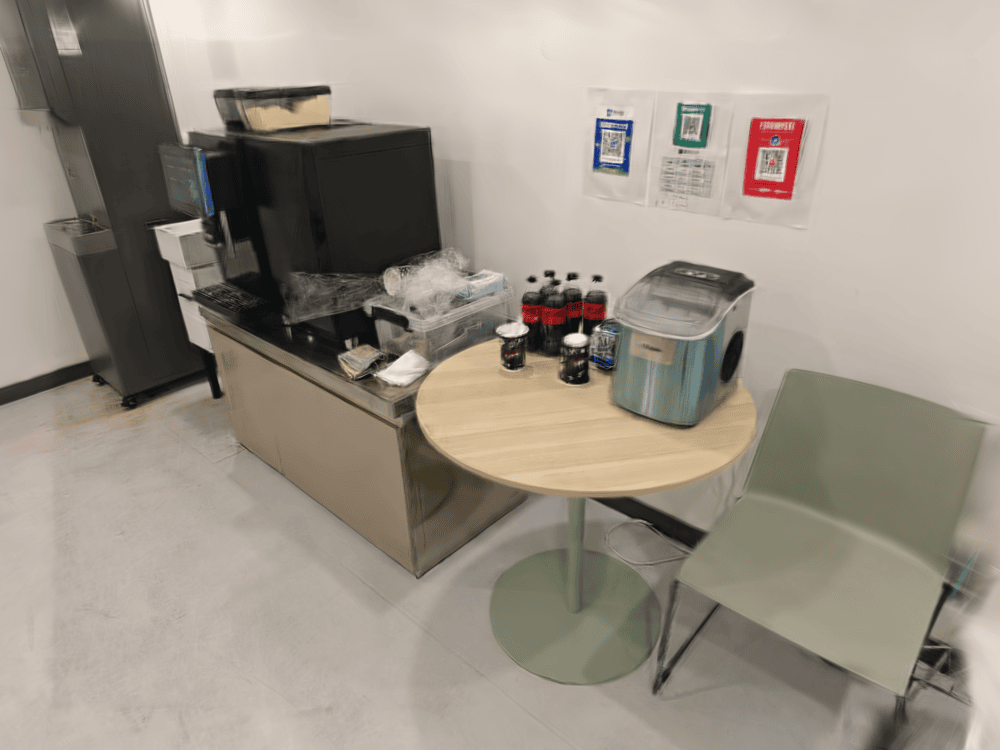} \\[2pt]
\rotatebox{90}{\,\,\scriptsize Factory bench} &
\includegraphics[width=0.18\linewidth]{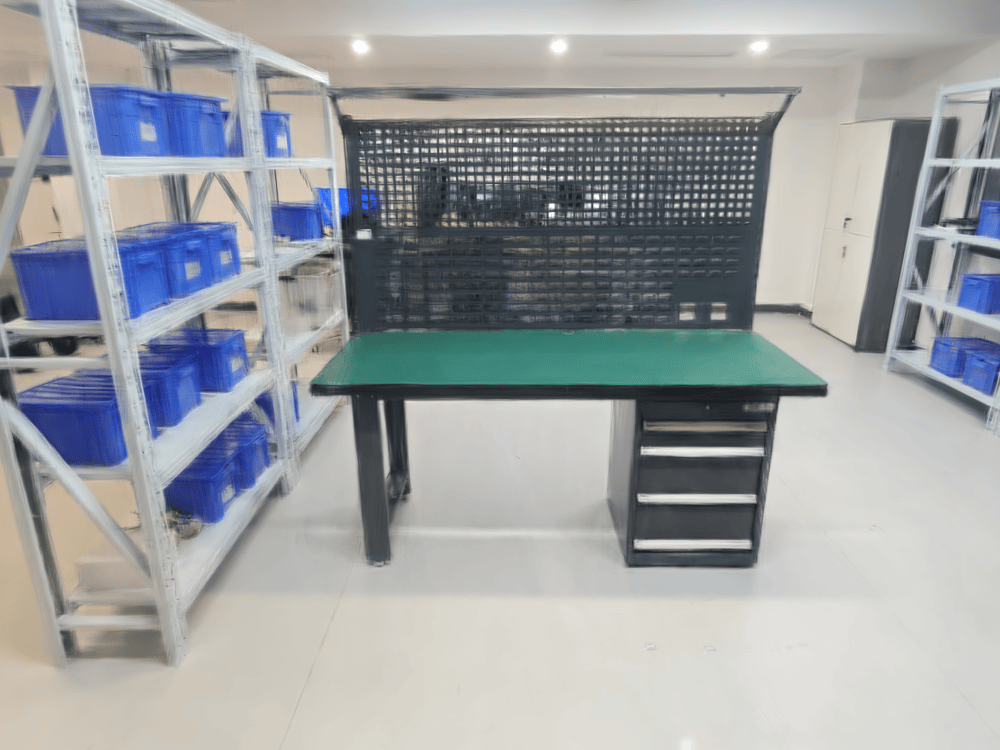} &
\includegraphics[width=0.18\linewidth]{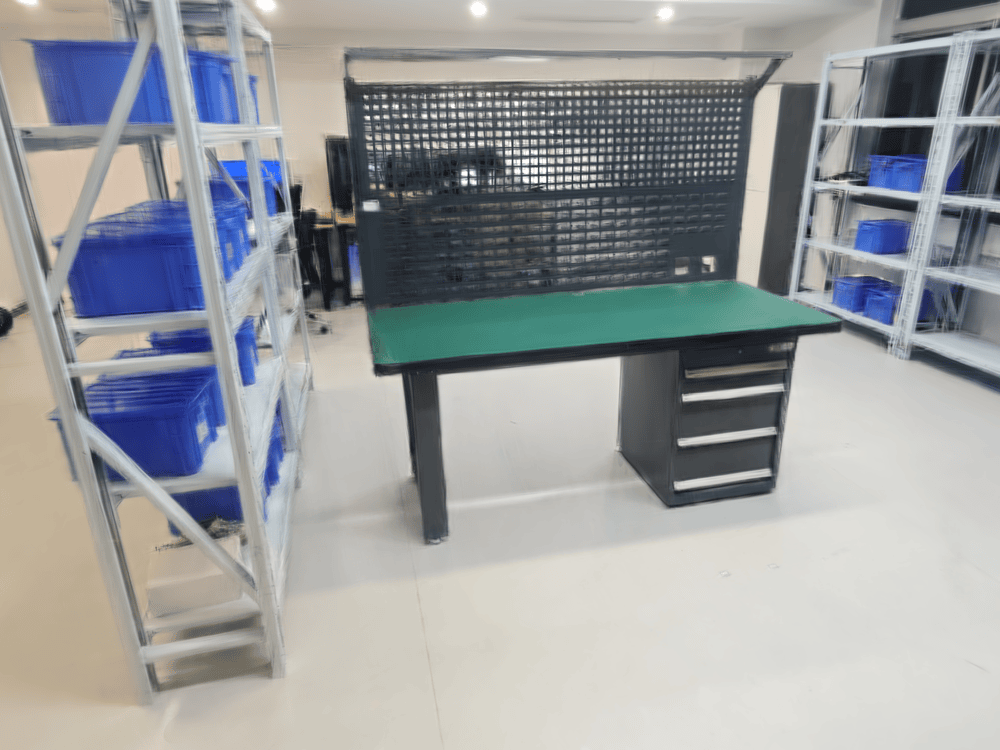} &
\includegraphics[width=0.18\linewidth]{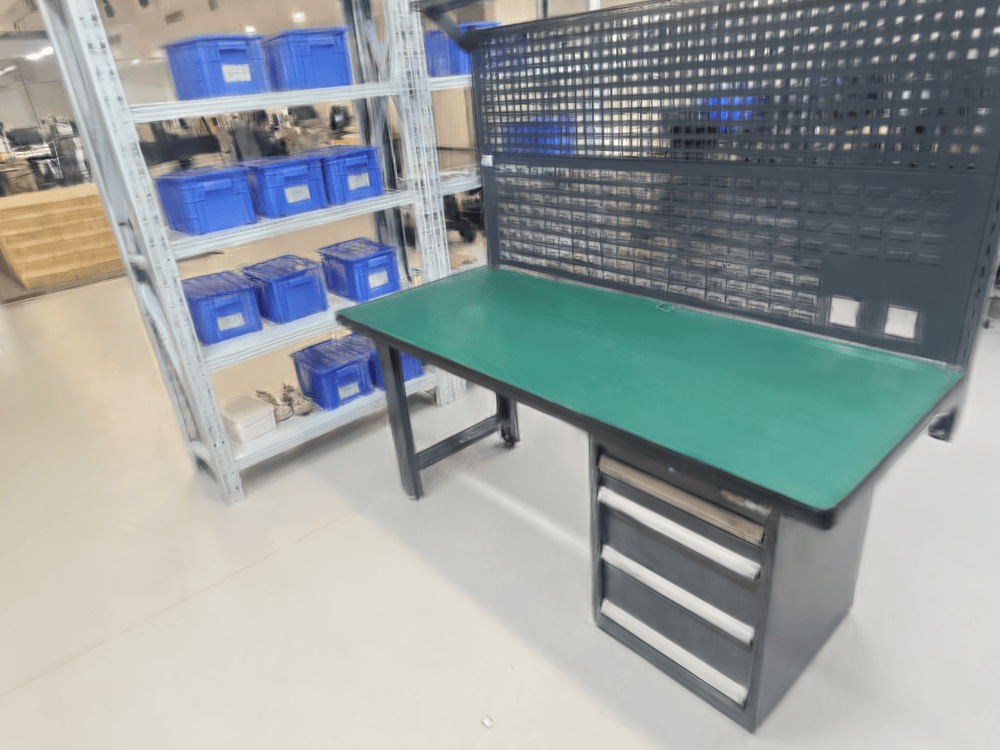} &
\includegraphics[width=0.18\linewidth]{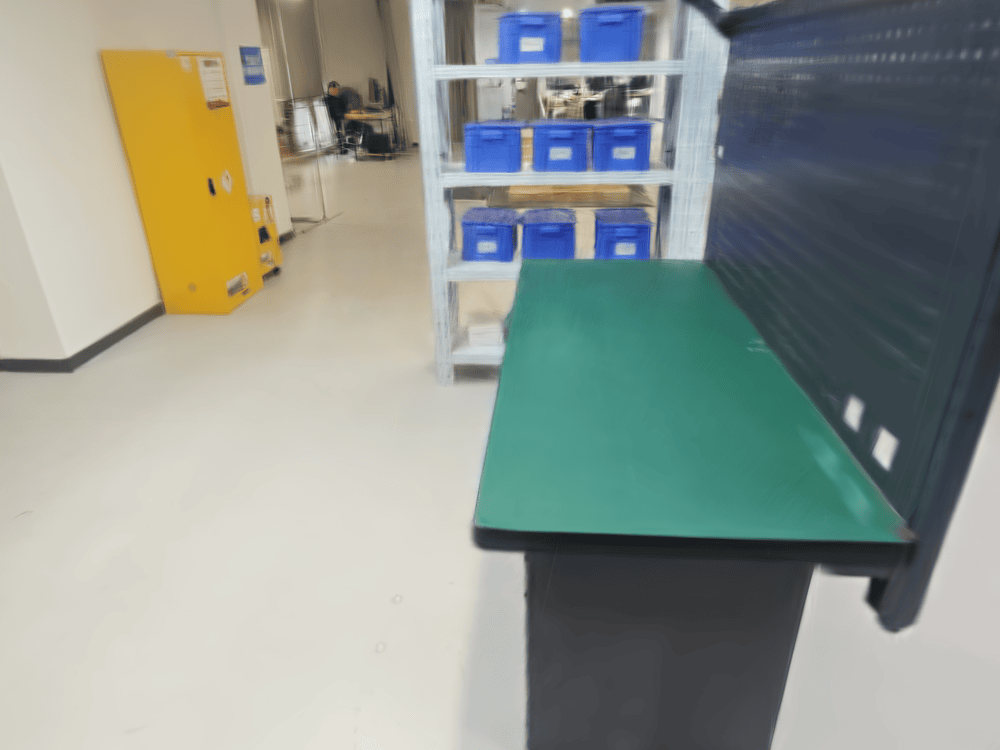} &
\includegraphics[width=0.18\linewidth]{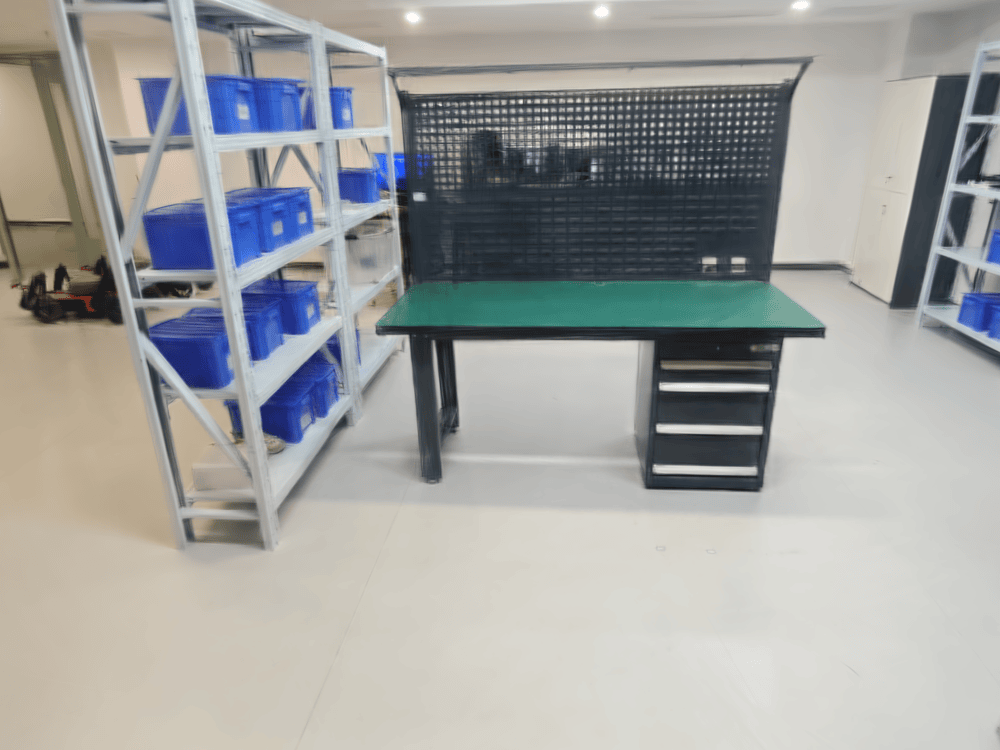} \\[2pt]
\rotatebox{90}{\,\,\scriptsize Library desk} &
\includegraphics[width=0.18\linewidth]{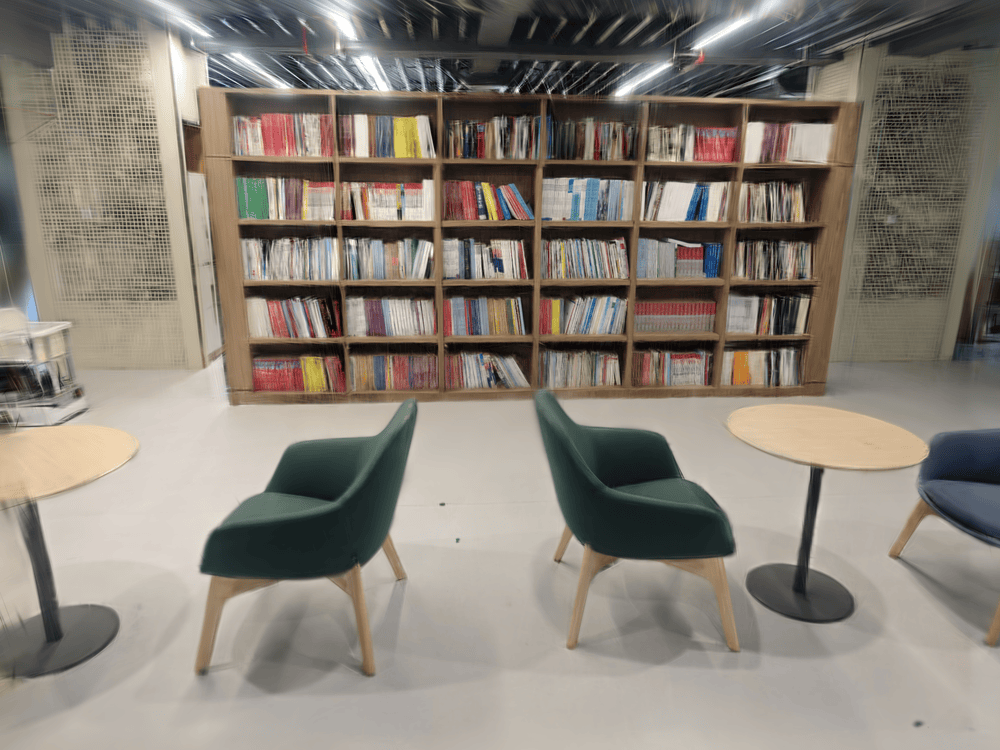} &
\includegraphics[width=0.18\linewidth]{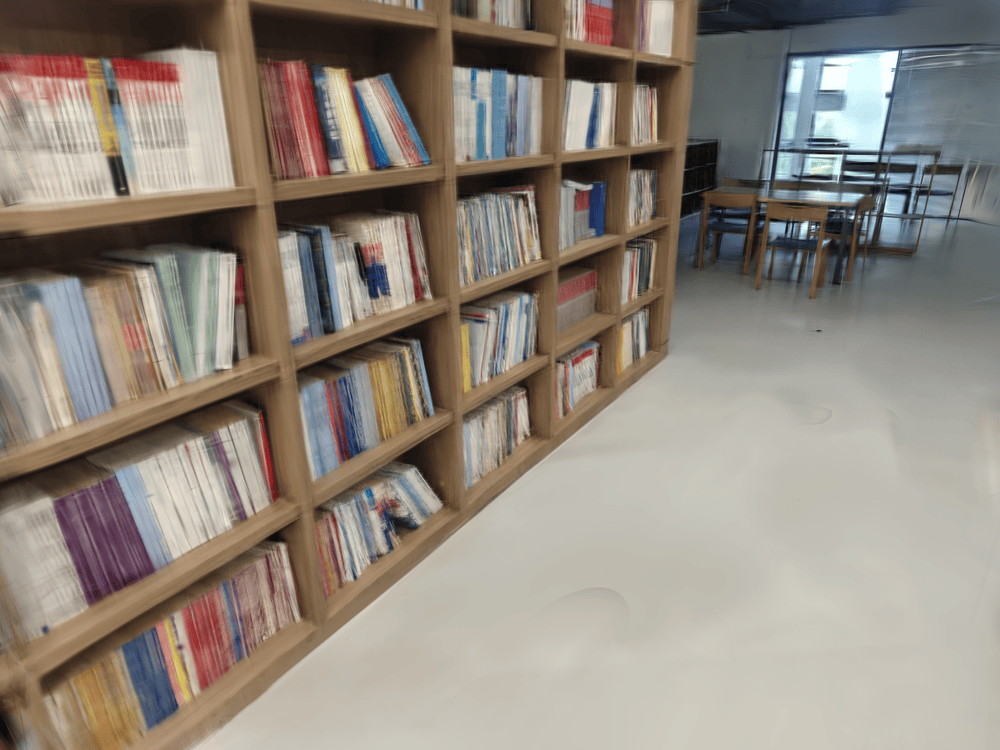} &
\includegraphics[width=0.18\linewidth]{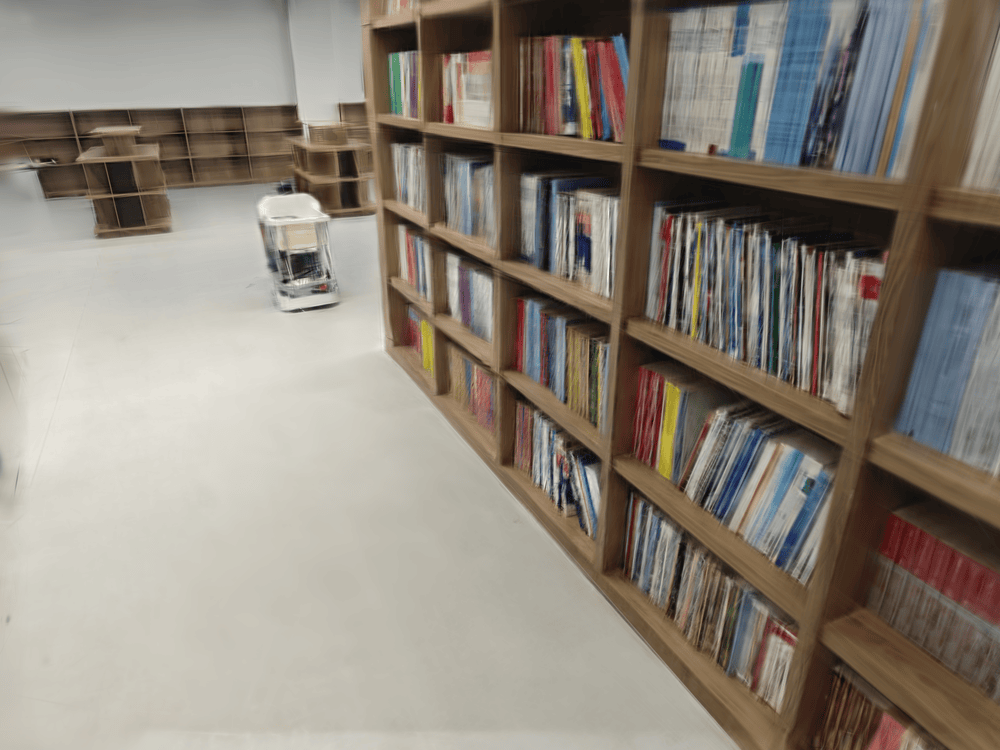} &
\includegraphics[width=0.18\linewidth]{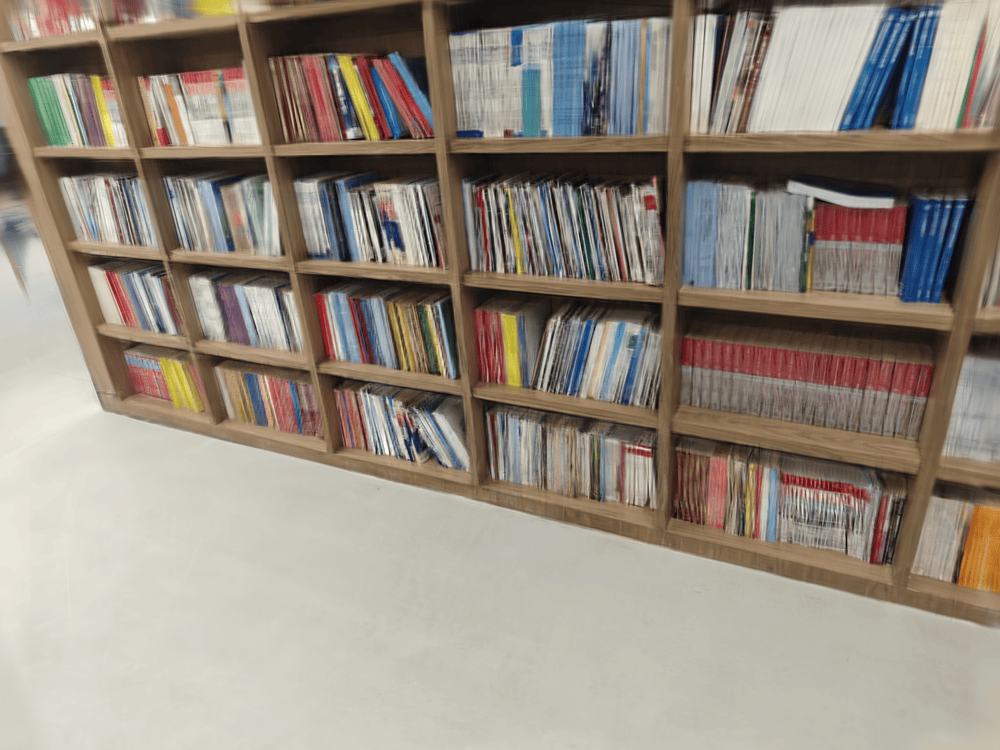} &
\includegraphics[width=0.18\linewidth]{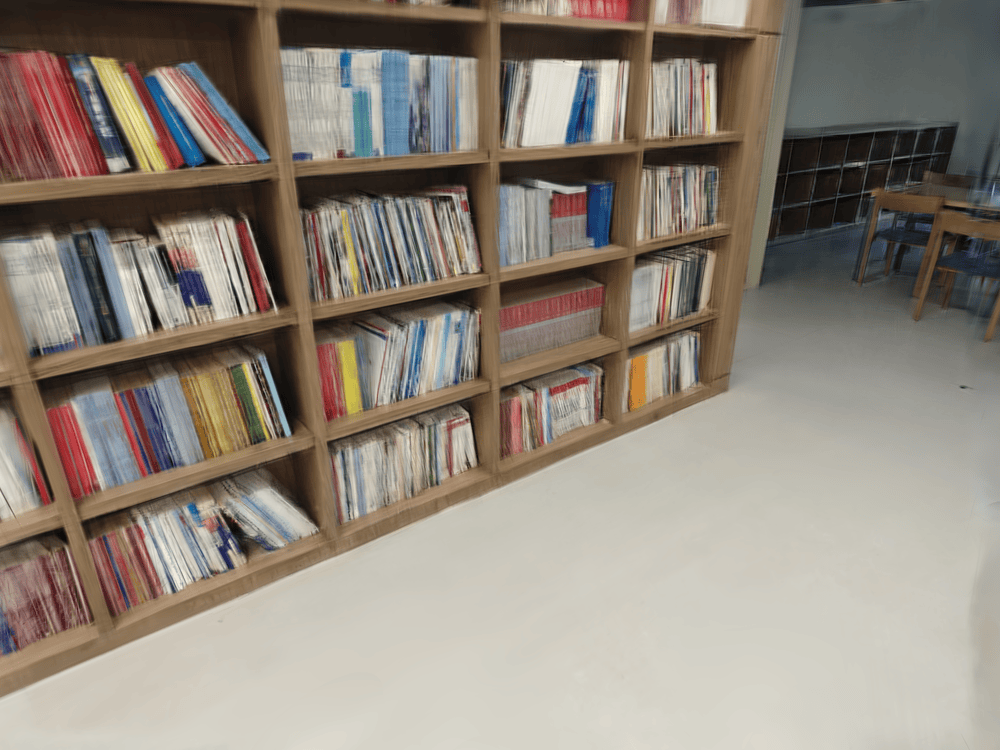} \\
\end{tabular}
\caption{\textbf{Depth Anything 3 input photographs.} Five representative phone photographs of each deployment scene, drawn from a set of 14--20 captures per scene. Each row corresponds to one of the three scenes reconstructed in Figure~\ref{fig:da3_recon}. The same photograph set is fed to Depth Anything 3 for joint depth and pose recovery, after which the 3D Gaussian Splatting field is fit and rendered into the simulator.}
\label{fig:da3_inputs}
\vspace{-0.5em}
\end{figure}
}

\newcommand{\figDAThreeRecon}{%
\begin{figure}[!htbp]
\centering
\setlength{\tabcolsep}{1.5pt}
\renewcommand{\arraystretch}{0.5}
\begin{tabular}{@{}cccccc@{}}
& {\scriptsize Canonical} & \multicolumn{4}{c}{\scriptsize Orbit renders at four representative angles} \\[1pt]
\rotatebox{90}{\,\,\scriptsize Coffee corner} &
\includegraphics[width=0.18\linewidth]{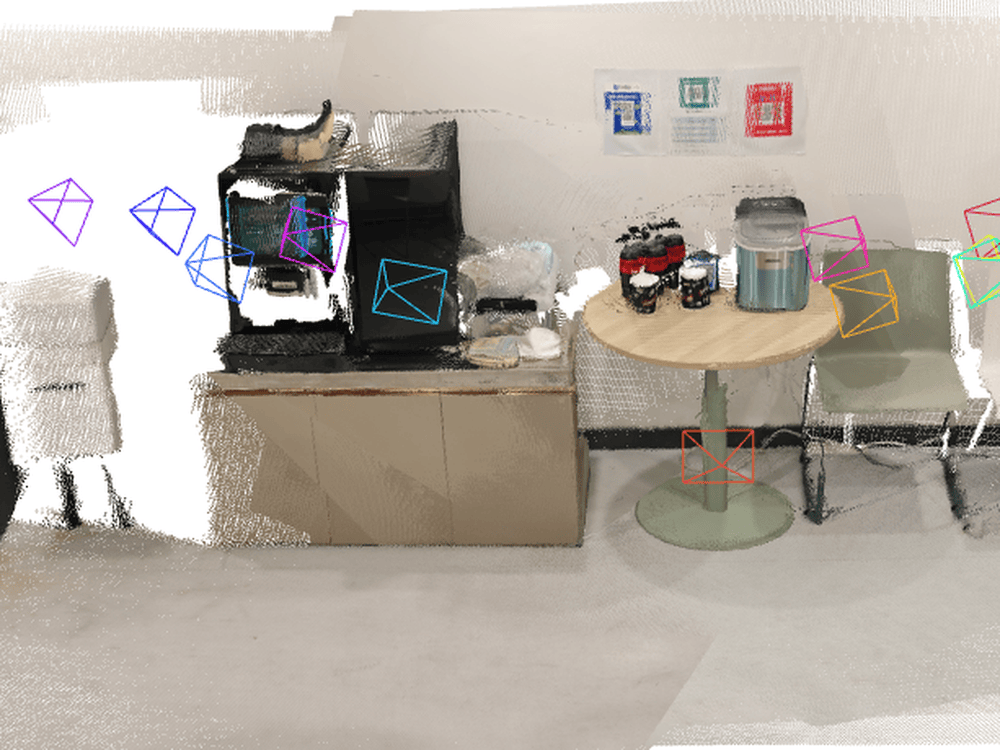} &
\includegraphics[width=0.18\linewidth]{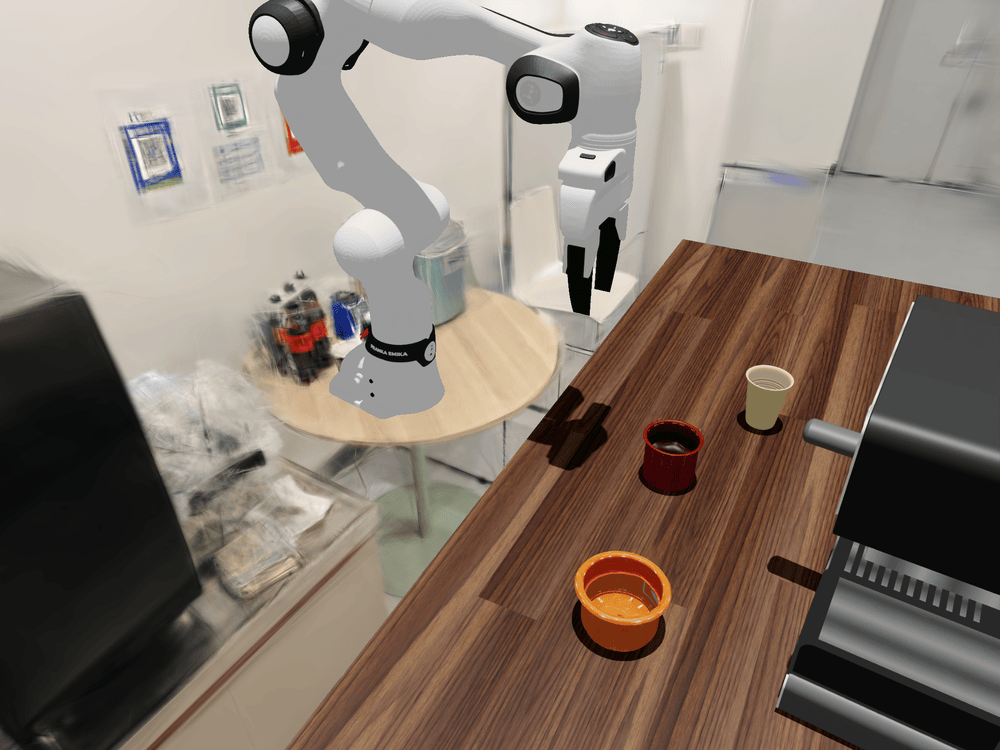} &
\includegraphics[width=0.18\linewidth]{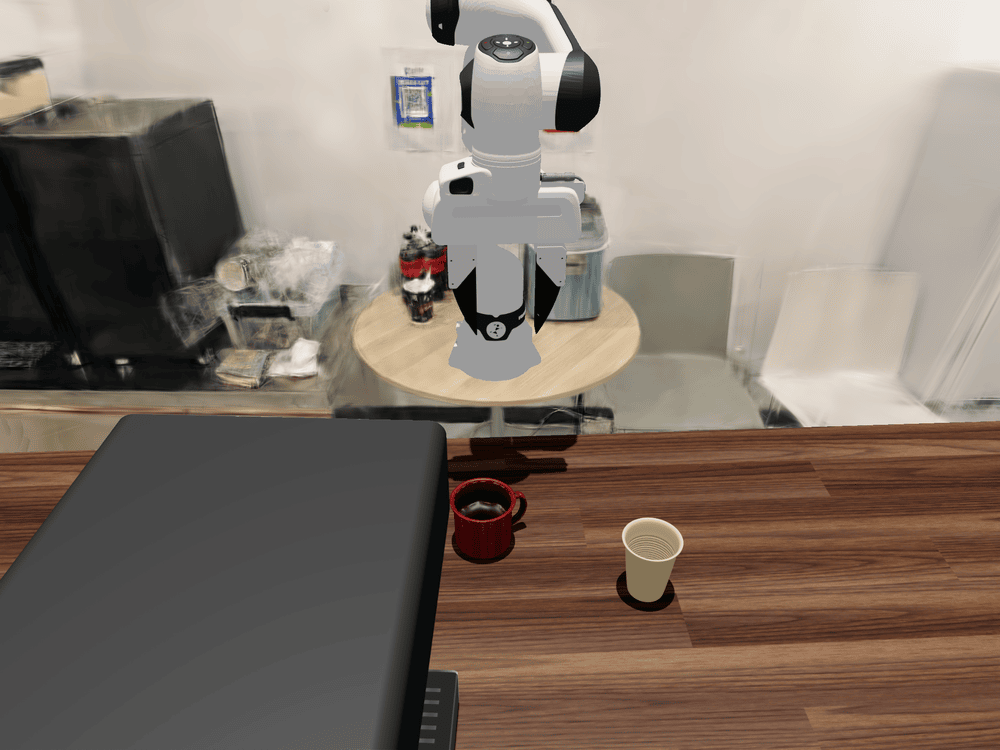} &
\includegraphics[width=0.18\linewidth]{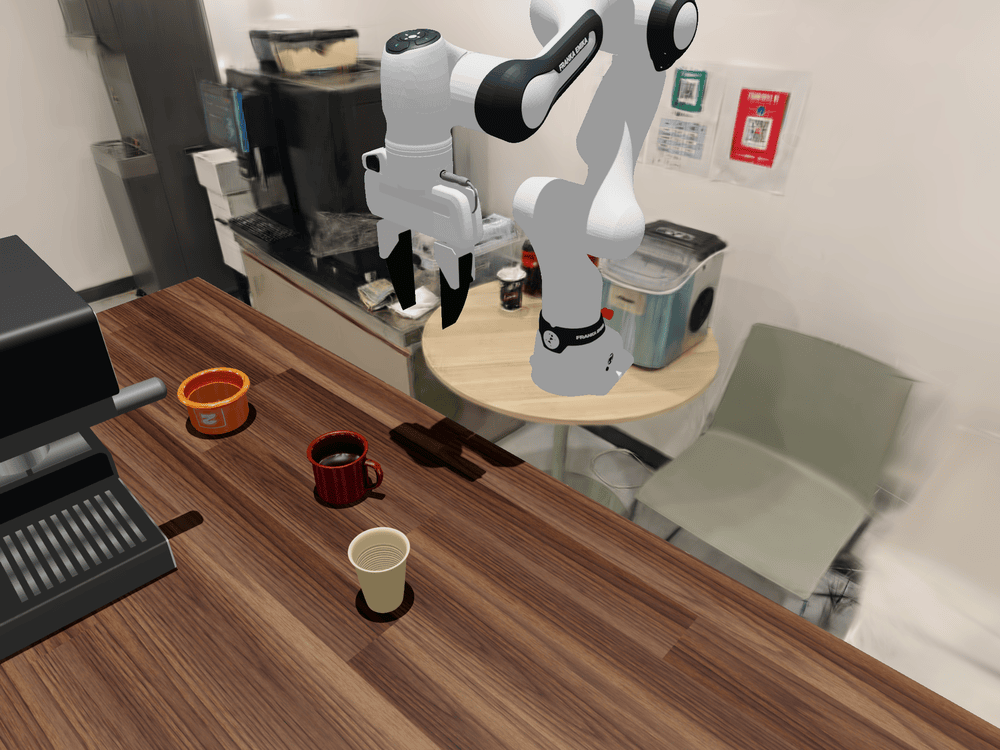} &
\includegraphics[width=0.18\linewidth]{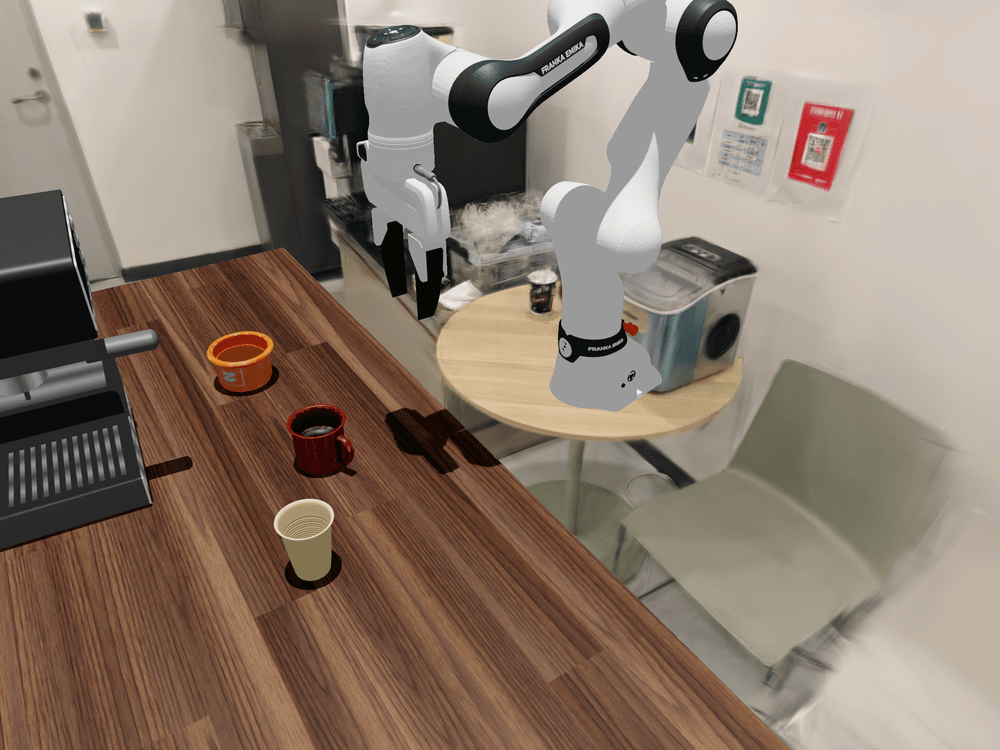} \\
& & {\scriptsize $-105^\circ$} & {\scriptsize $-45^\circ$} & {\scriptsize $0^\circ$} & {\scriptsize $+15^\circ$} \\[2pt]
\rotatebox{90}{\,\,\scriptsize Factory bench} &
\includegraphics[width=0.18\linewidth]{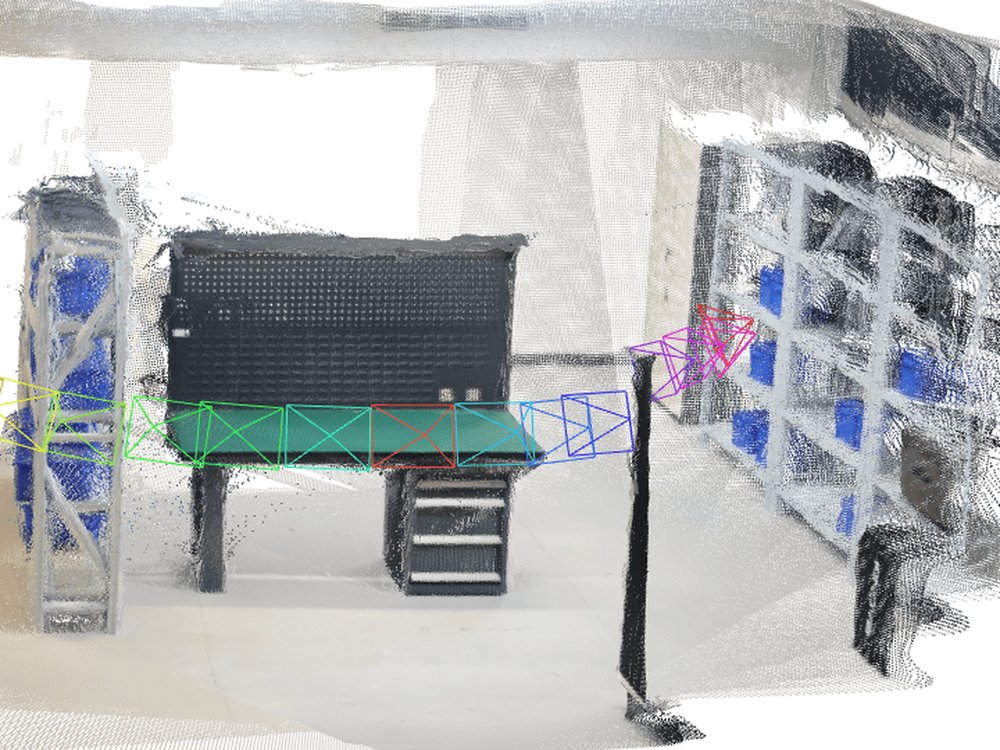} &
\includegraphics[width=0.18\linewidth]{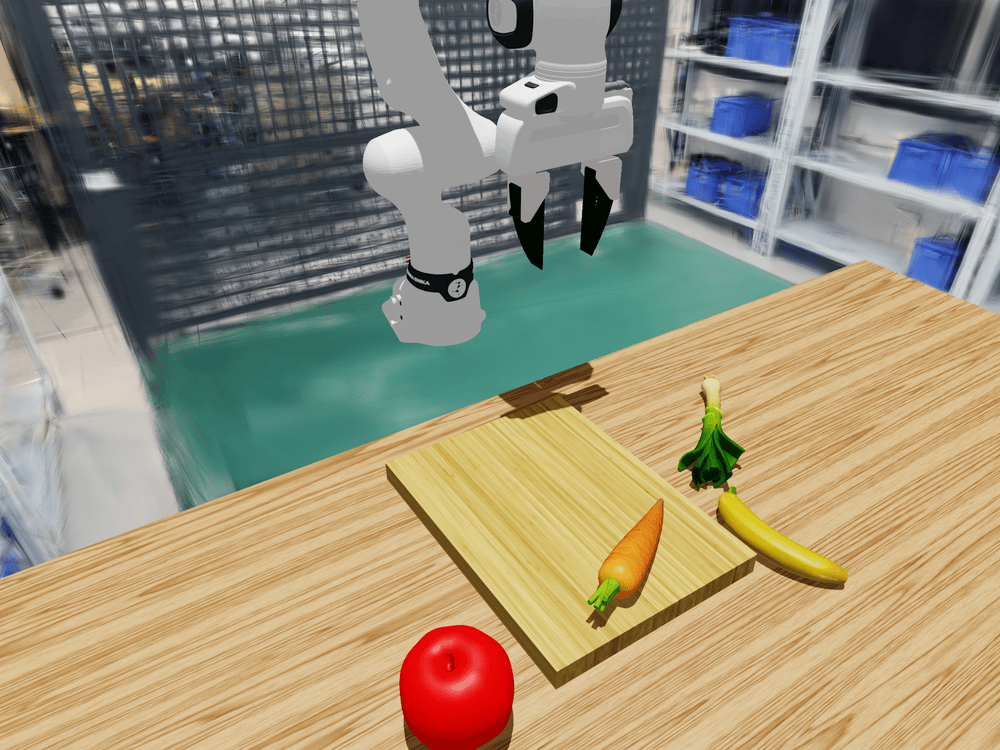} &
\includegraphics[width=0.18\linewidth]{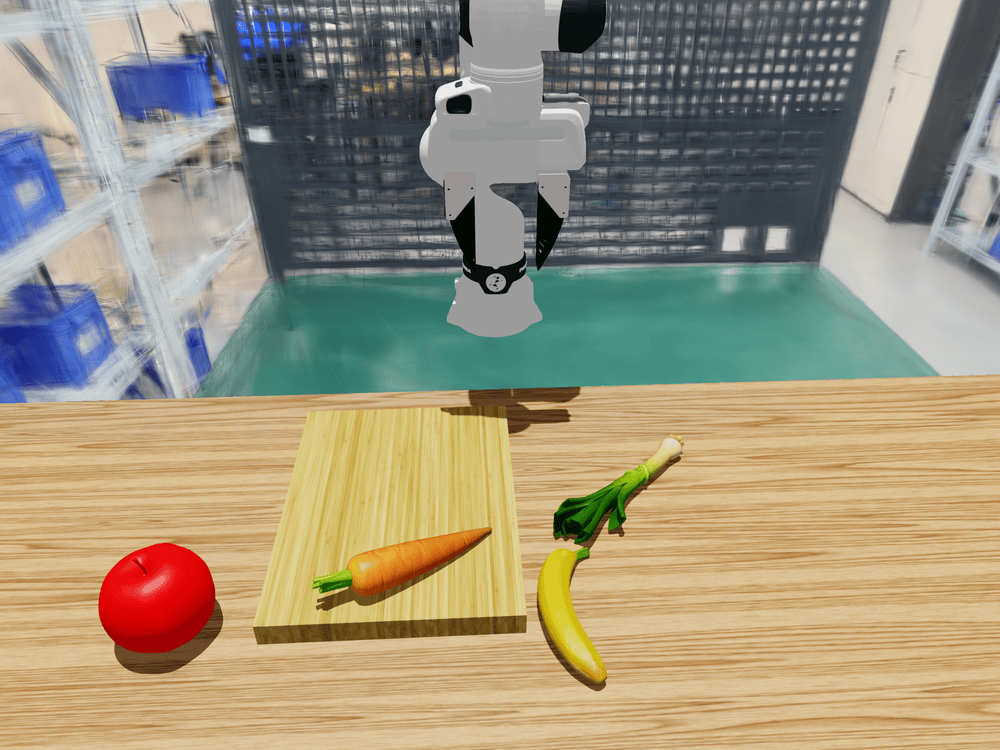} &
\includegraphics[width=0.18\linewidth]{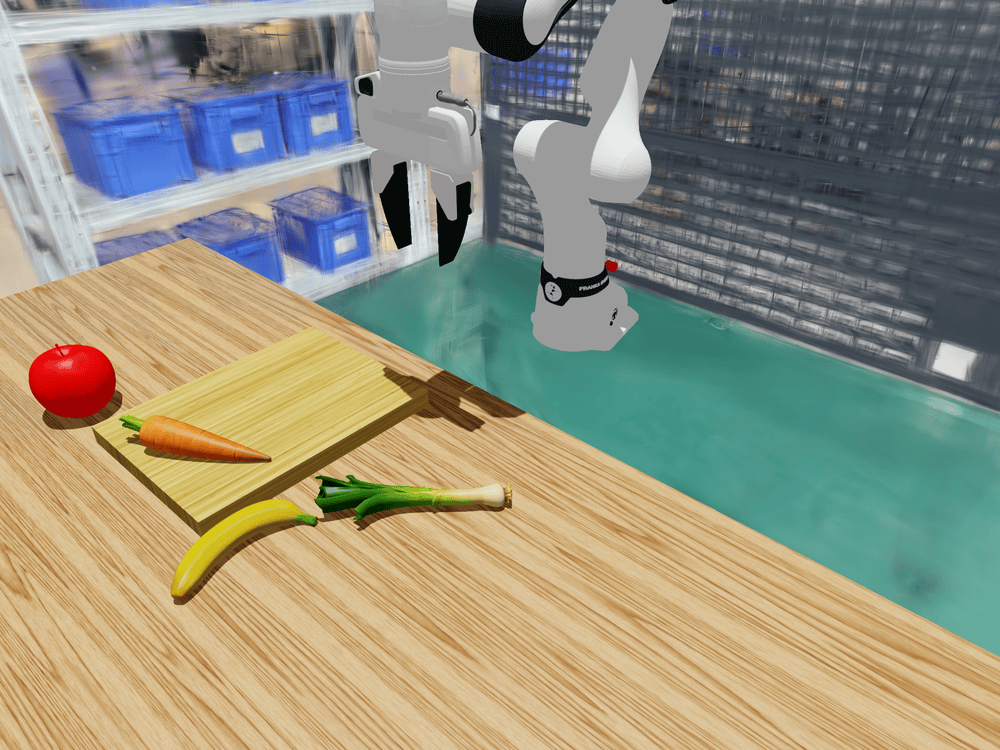} &
\includegraphics[width=0.18\linewidth]{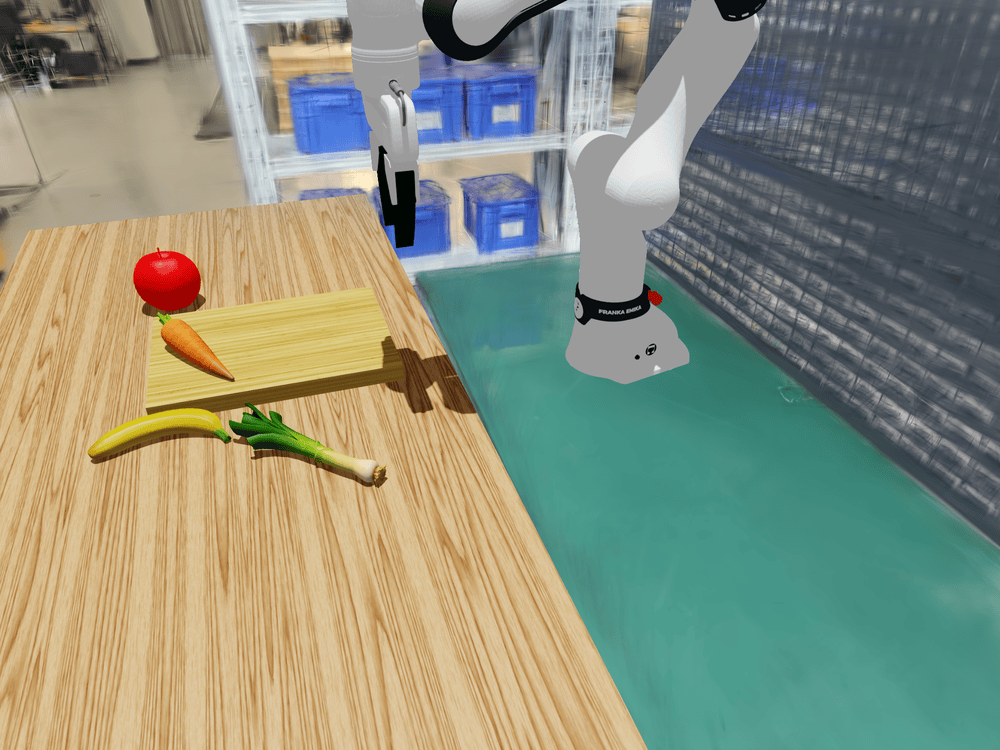} \\
& & {\scriptsize $-75^\circ$} & {\scriptsize $-45^\circ$} & {\scriptsize $0^\circ$} & {\scriptsize $+30^\circ$} \\[2pt]
\rotatebox{90}{\,\,\scriptsize Library desk} &
\includegraphics[width=0.18\linewidth]{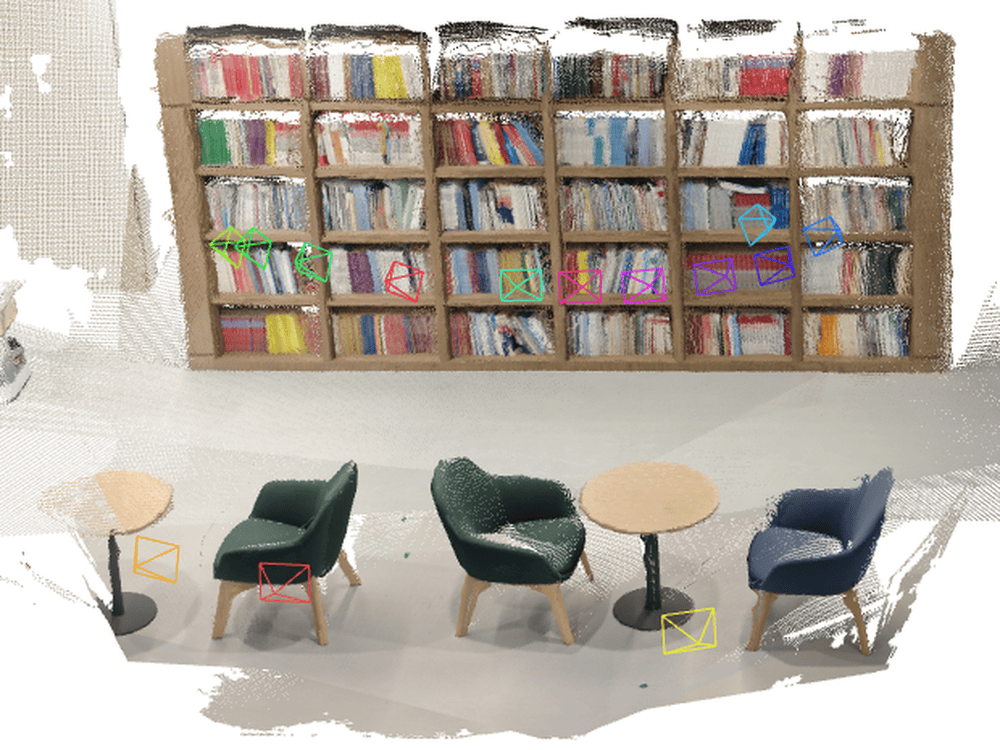} &
\includegraphics[width=0.18\linewidth]{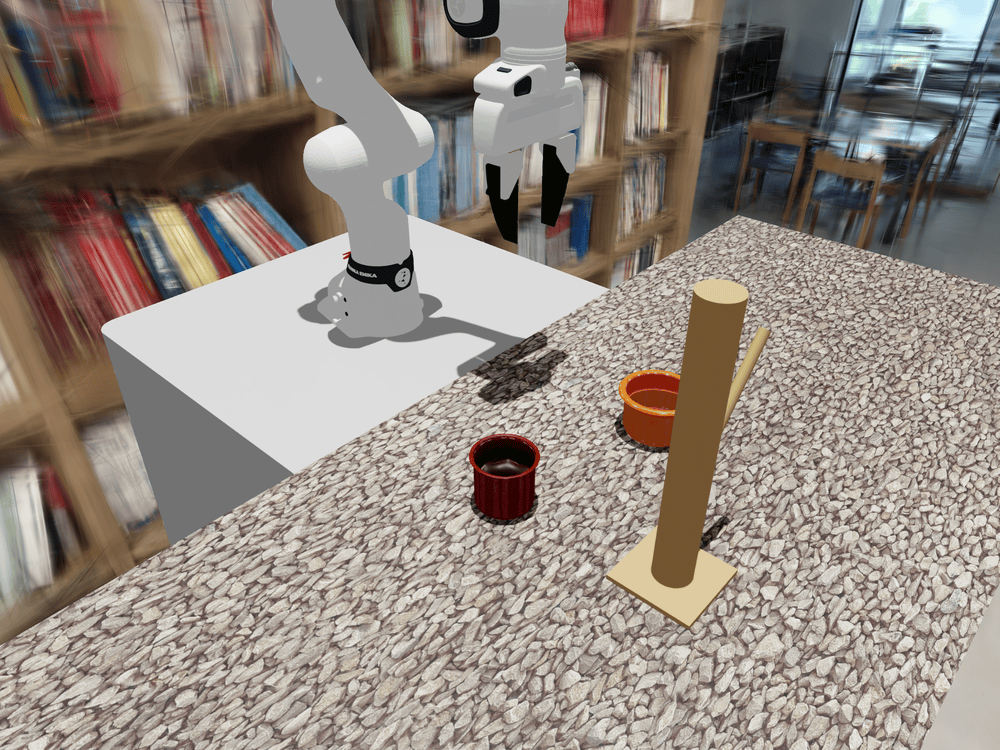} &
\includegraphics[width=0.18\linewidth]{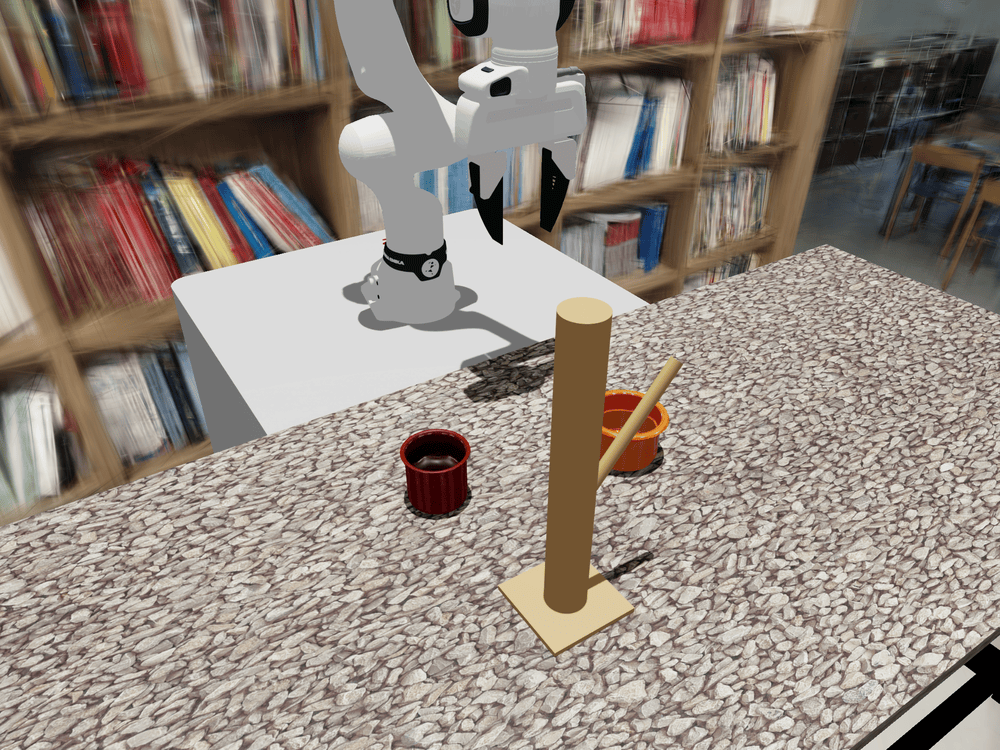} &
\includegraphics[width=0.18\linewidth]{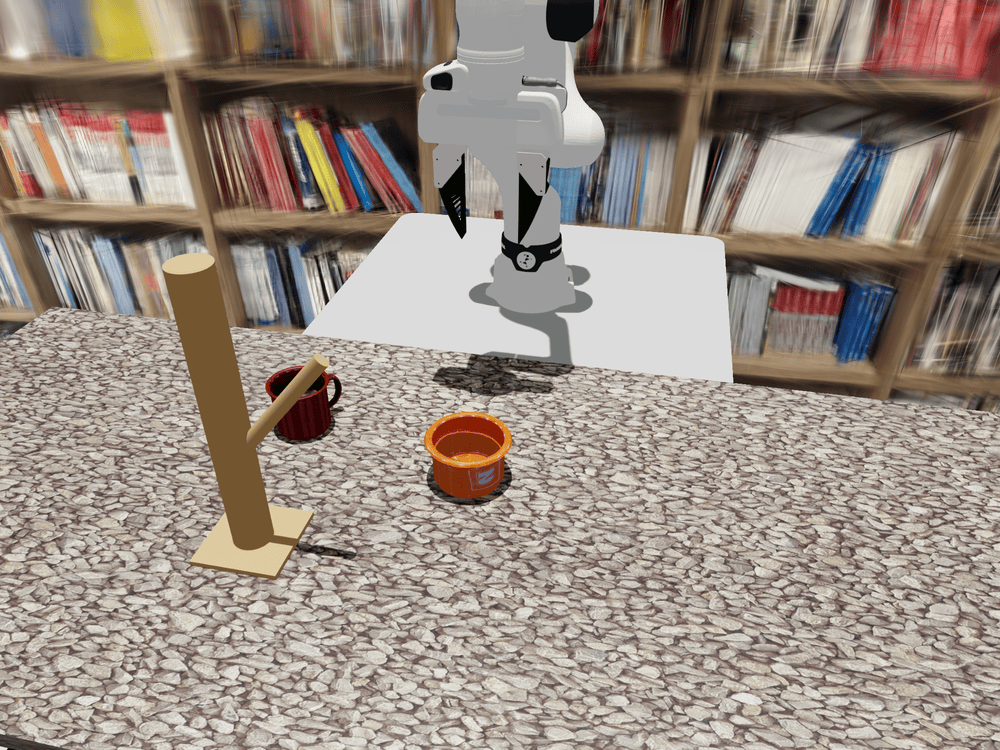} &
\includegraphics[width=0.18\linewidth]{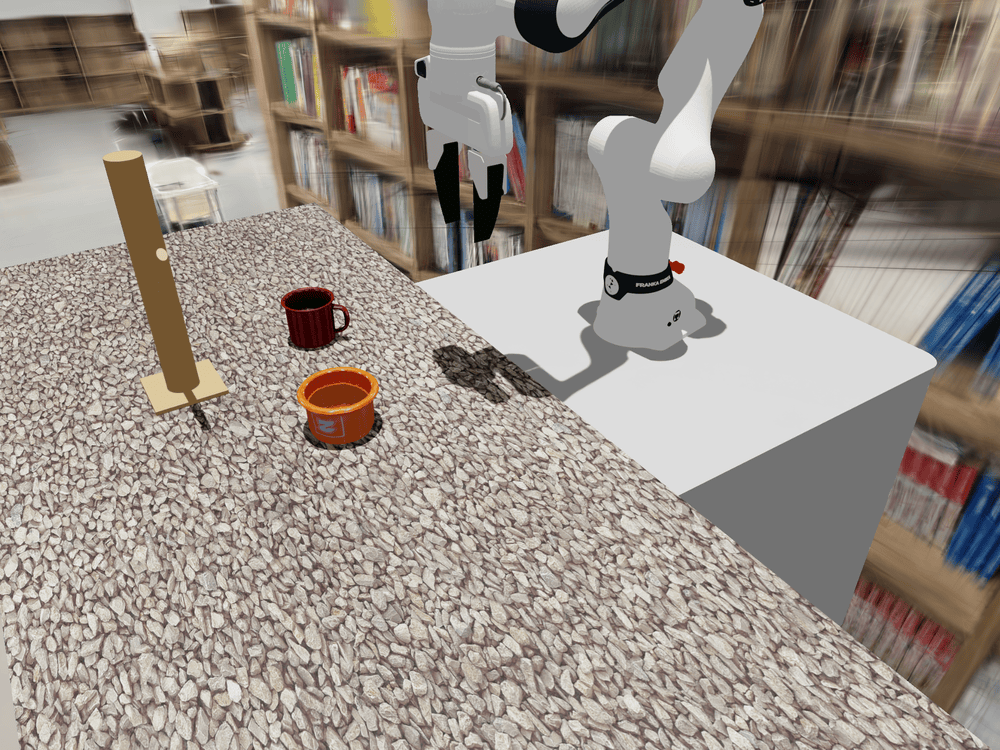} \\
& & {\scriptsize $-90^\circ$} & {\scriptsize $-75^\circ$} & {\scriptsize $-30^\circ$} & {\scriptsize $+15^\circ$} \\
\end{tabular}
\caption{\textbf{Depth Anything 3 background reconstruction across three deployment scenes.} Each row is one real workspace reconstructed from 10--20 photographs by the pipeline described in Appendix~\ref{app:da3}. The first column shows the canonical viewpoint render and the next four columns show novel viewpoints generated by orbiting a virtual camera around the workspace, at four representative angles within the supported $[-120^\circ,+45^\circ]$ orbit range.}
\label{fig:da3_recon}
\vspace{-0.5em}
\end{figure}
}

\newcommand{\tableBenchmark}{%
\begin{table*}[!htbp]
\caption{\textbf{Benchmark evaluation results} (Success Rate \%) on 20 representative tasks organized by the three difficulty tiers of Section~\ref{sec:benchmark}. Each task is evaluated over 100 rollouts. Best result per task in \textbf{bold}; tier averages bolded.}
\label{tab:benchmark}
\centering
\small
\setlength{\tabcolsep}{4pt}
\begin{tabular}{llccccc}
\toprule
\textbf{Tier} & \multicolumn{1}{c}{\textbf{Task}} & \textbf{BC} & \textbf{DP} & \textbf{ACT} & \textbf{VLA-adapter} & \textbf{Pi\,0.5} \\
\midrule
\multirow{4}{*}{Easy}
 & \textit{pick banana place plate}                   & 40 & 82 & 47 & 87 & \textbf{93} \\
 & \textit{pick leek place cuttingboard}              & 33 & 70 & 68 & 84 & \textbf{86} \\
 & \textit{pick chocolatebox place sugarbox}          & 43 & 76 & 79 & 88 & \textbf{89} \\
 & \multicolumn{1}{l}{\textbf{Tier average (3 tasks)}} & \textbf{39} & \textbf{76} & \textbf{65} & \textbf{86} & \textbf{89} \\
\midrule
\multirow{13}{*}{Medium}
 & \textit{pick cup place shelf}                      &  7 & 28 & 26 & 55 & \textbf{61} \\
 & \textit{pick mug place coffee machine}             & 10 & 33 & 27 & 51 & \textbf{53} \\
 & \textit{pick lid place pot}                        & 13 & 37 & 17 & \textbf{53} & 50 \\
 & \textit{pick soupcan place microwave}              & 21 & 45 & 41 & 67 & \textbf{69} \\
 & \textit{pour scoop into bowl}                      & 13 & 35 & 31 & 62 & \textbf{67} \\
 & \textit{pour red mug into bowl}                    &  3 & 37 & 30 & 62 & \textbf{68} \\
 & \textit{pour kettle into bowl}                     &  3 & 33 & 26 & \textbf{56} & 53 \\
 & \textit{stack three cubes}                         &  4 & 38 & 36 & \textbf{57} & 47 \\
 & \textit{hang white cup on rack}                    &  4 & 23 & 10 & \textbf{47} & 42 \\
 & \textit{push can to target}                        & 27 & 57 & 53 & 70 & \textbf{73} \\
 & \textit{pull cabinet open}                         & 24 & 60 & 57 & \textbf{74} & 71 \\
 & \textit{open microwave}                            & 20 & 38 & 40 & \textbf{60} & 57 \\
 & \multicolumn{1}{l}{\textbf{Tier average (12 tasks)}} & \textbf{12} & \textbf{39} & \textbf{33} & \textbf{60} & \textbf{59} \\
\midrule
\multirow{6}{*}{Hard}
 & \textit{hang red mug on rack}                      &  3 & 23 & 21 & \textbf{36} & 34 \\
 & \textit{pour kettle and teapot}                    &  0 &  4 &  2 & 18 & \textbf{21} \\
 & \textit{store cups on plates}                      &  0 &  4 &  0 & 10 & \textbf{12} \\
 & \textit{pour sugar pick mug into coffeemachine}    &  0 &  4 &  4 & 18 & \textbf{20} \\
 & \textit{organize the clutter into baskets}         &  0 & 15 &  9 & 32 & \textbf{35} \\
 & \multicolumn{1}{l}{\textbf{Tier average (5 tasks)}} & \textbf{1} & \textbf{10} & \textbf{7} & \textbf{23} & \textbf{24} \\
\midrule
\multicolumn{2}{c}{\textbf{Overall (20 tasks)}} & \textbf{13} & \textbf{37} & \textbf{31} & \textbf{54} & \textbf{55} \\
\bottomrule
\end{tabular}
\vspace{-1em}
\end{table*}
}

\newcommand{\tableAffordMain}{%
\begin{table}[!htb]
\caption{\textbf{Affordance integration on the AffordSim benchmark.} Trajectory collection success rate (\%) per task across the five grasp methods. AnyGrasp+Filter denotes AnyGrasp candidates filtered against a 2D SAM2 task-relevant mask, mirroring the GenManip-style affordance-filtering pipeline. Two representative tasks per tier are listed; full breakdown in Appendix~\ref{app:afford_full}. Tier averages bolded.}
\label{tab:afford_main}
\centering
\footnotesize
\setlength{\tabcolsep}{4pt}
\begin{tabular}{llccccc}
\toprule
\textbf{Tier} & \multicolumn{1}{c}{\textbf{Task}} & \textbf{Manual} & \textbf{AnyGrasp} & \textbf{AnyGrasp+Filter} & \textbf{AffordSim} & \textbf{Human Aff.} \\
\midrule
\multirow{4}{*}{Easy}
  & \textit{pick banana place bowl}                   & 100 & 80 & 100 & 100 & 100 \\
  & \textit{pick soupcan place table}                 & 100 & 55 &  75 &  95 & 100 \\
  & \multicolumn{1}{l}{$\cdots$ (8 more tasks)}       & $\cdots$ & $\cdots$ & $\cdots$ & $\cdots$ & $\cdots$ \\
  & \textbf{Average (10 tasks)}                       & \textbf{100} & \textbf{67} & \textbf{87}  & \textbf{98}  & \textbf{100} \\
\midrule
\multirow{4}{*}{Medium}
  & \textit{pick mugcup place coffeemachine}          & 90 & 25 & 55 & 88 & 95 \\
  & \textit{hang cup}                                 & 85 &  0 &  0 & 25 & 90 \\
  & \multicolumn{1}{l}{$\cdots$ (28 more tasks)}      & $\cdots$ & $\cdots$ & $\cdots$ & $\cdots$ & $\cdots$ \\
  & \textbf{Average (30 tasks)}                       & \textbf{85}  & \textbf{15} & \textbf{45} & \textbf{79}  & \textbf{93}  \\
\midrule
\multirow{4}{*}{Hard}
  & \textit{stack four cube}                          & 82 & 18 & 33 & 77 & 88 \\
  & \textit{composite double pour kettle}             & 70 &  0 & 15 & 65 & 85 \\
  & \multicolumn{1}{l}{$\cdots$ (8 more tasks)}       & $\cdots$ & $\cdots$ & $\cdots$ & $\cdots$ & $\cdots$ \\
  & \textbf{Average (10 tasks)}                       & \textbf{72}  & \textbf{3}  & \textbf{16} & \textbf{64}  & \textbf{86}  \\
\bottomrule
\end{tabular}
\vspace{-1em}
\end{table}
}

\newcommand{\figTabAffordSimReal}{%
\begin{figure}[H]
\centering
\begin{minipage}[t]{0.42\textwidth}
\vspace{0pt}
\centering
\includegraphics[width=\linewidth]{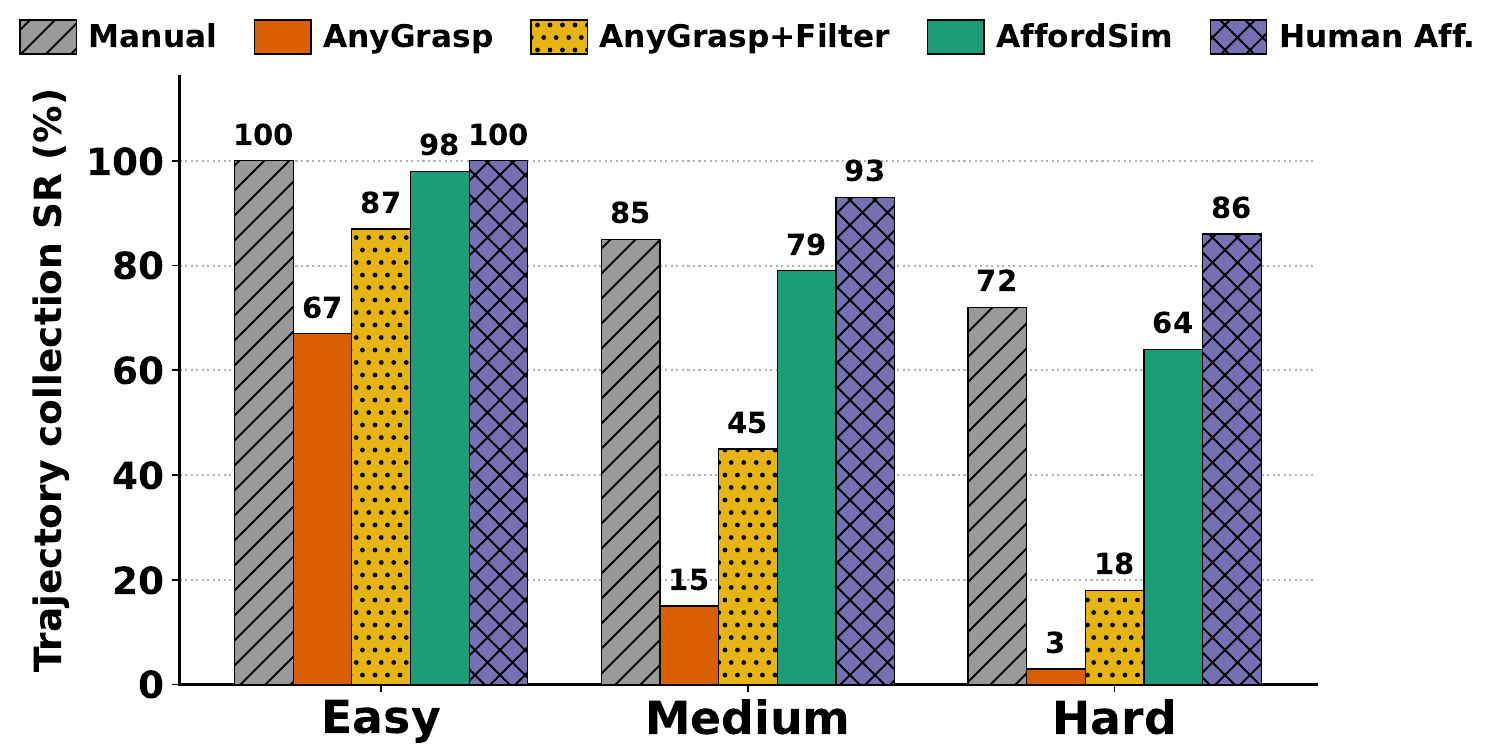}
\captionsetup{width=\linewidth}
\caption{\textbf{Per-tier average trajectory collection SR (\%).}}
\label{fig:afford_tier}
\end{minipage}\hfill
\begin{minipage}[t]{0.56\textwidth}
\vspace{0pt}
\centering
\captionsetup{width=\linewidth}
\captionof{table}{\textbf{Zero-shot sim-to-real transfer.} SR (\%) on a real Franka FR3 (10 trials per task, no real world fine tuning). Best per task in \textbf{bold}.}
\label{tab:sim2real}
\scriptsize
\setlength{\tabcolsep}{2.5pt}
\begin{tabular}{lccccc}
\toprule
\textbf{Task} & \textbf{BC} & \textbf{DP} & \textbf{ACT} & \textbf{VLA-A.} & \textbf{Pi\,0.5} \\
\midrule
\textit{pick banana place plate} & 0 & 20 & 10 & \textbf{40} & \textbf{40} \\
\textit{pick cup place shelf}    & 0 &  0 &  0 & \textbf{20} & \textbf{20} \\
\textit{pour scoop into bowl}    & 0 & 10 &  0 & 20          & \textbf{30} \\
\textit{hang red mug on rack}    & 0 &  0 &  0 & \textbf{10} & \textbf{10} \\
\midrule
\textbf{Average} & \textbf{0} & \textbf{8} & \textbf{3} & \textbf{23} & \textbf{25} \\
\bottomrule
\end{tabular}
\end{minipage}
\vspace{-1em}
\end{figure}
}

\newcommand{\tableAffordFull}{%
\begin{table}[!htbp]
\caption{\textbf{Full per-task AffordSim trajectory collection success rate.} Per-task success rate (\%) for every task of the AffordSim benchmark across the five grasp methods, grouped by difficulty tier as in Table~\ref{tab:afford_main}. Each task is evaluated over 100 rollouts. Per-tier averages are bolded.}
\label{tab:afford_full}
\centering
\footnotesize
\setlength{\tabcolsep}{4.5pt}
\renewcommand{\arraystretch}{0.88}
\begin{tabular}{llccccc}
\toprule
\textbf{Tier} & \multicolumn{1}{c}{\textbf{Task}} & \textbf{Manual} & \textbf{AnyGrasp} & \textbf{AnyGrasp+Filter} & \textbf{AffordSim} & \textbf{Human Aff.} \\
\midrule
\multirow{10}{*}{Easy}
  & \textit{pick banana place bowl}              & 100 & 80 & 100 & 100 & 100 \\
  & \textit{pick bottle place table}             & 100 & 70 &  90 & 100 & 100 \\
  & \textit{pick bowl place plate}               & 100 & 67 &  87 & 100 & 100 \\
  & \textit{pick carrot place bowl}              & 100 & 73 &  90 & 100 & 100 \\
  & \textit{pick carrot place plate}             & 100 & 67 &  87 & 100 & 100 \\
  & \textit{pick chocolatebox place sugarbox}    & 100 & 70 &  87 &  93 & 100 \\
  & \textit{pick cube place cup}                 & 100 & 73 &  87 & 100 & 100 \\
  & \textit{pick leek place cuttingboard}        & 100 & 67 &  87 & 100 & 100 \\
  & \textit{pick pen place doraemoncup}          & 100 & 60 &  80 &  90 & 100 \\
  & \textit{pick soupcan place table}            & 100 & 55 &  75 &  95 & 100 \\
\cmidrule{2-7}
\multicolumn{2}{l}{\textbf{Easy average (10 tasks)}}        & \textbf{100} & \textbf{67} & \textbf{87}  & \textbf{98}  & \textbf{100} \\
\midrule
\multirow{30}{*}{Medium}
  & \textit{close microwave}                     &  70 &  5 &  22 &  65 &  73 \\
  & \textit{composite 2picks to plate}           &  80 &  8 &  30 &  72 &  90 \\
  & \textit{composite 3picks to paperbox}        &  70 &  5 &  25 &  60 &  83 \\
  & \textit{hang cup}                            &  85 &  0 &   0 &  25 &  90 \\
  & \textit{open microwave}                      &  78 &  8 &  28 &  72 &  82 \\
  & \textit{pick cup place coffeemachine}        &  87 & 18 &  47 &  82 & 100 \\
  & \textit{pick cup place woodshelf}            &  88 & 22 &  53 &  85 & 100 \\
  & \textit{pick green bellpepper place pan}     &  90 & 25 &  55 &  87 & 100 \\
  & \textit{pick kettle place base}              &  87 & 12 &  40 &  82 &  93 \\
  & \textit{pick lid place pot}                  &  85 &  8 &  33 &  78 & 100 \\
  & \textit{pick mugcup place coffeemachine}     &  90 & 25 &  55 &  88 &  95 \\
  & \textit{pick mugcup place woodshelf}         &  87 & 20 &  50 &  82 &  87 \\
  & \textit{pick papercup place table}           &  87 & 30 &  62 &  88 & 100 \\
  & \textit{pick scrubbottle place woodshelf}    &  85 & 22 &  55 &  85 & 100 \\
  & \textit{pick soupcan place microwave}        &  92 & 28 &  62 &  88 & 100 \\
  & \textit{pour basket}                         &  78 &  5 &  28 &  67 &  73 \\
  & \textit{pour bluekettle}                     &  87 & 18 &  55 &  88 & 100 \\
  & \textit{pour mugcup}                         &  85 & 10 &  38 &  78 &  87 \\
  & \textit{pour pan}                            &  75 &  5 &  25 &  65 &  67 \\
  & \textit{pour papercup}                       &  85 & 12 &  40 &  78 &  87 \\
  & \textit{pour whitekettle}                    &  88 & 18 &  47 &  87 &  93 \\
  & \textit{pour whitekettle4}                   &  87 & 15 &  45 &  83 &  87 \\
  & \textit{pull cabinet}                        &  90 & 18 &  58 &  90 & 100 \\
  & \textit{pull cabinet7}                       &  87 & 15 &  55 &  88 & 100 \\
  & \textit{pull cabinet15}                      &  88 & 12 &  50 &  87 & 100 \\
  & \textit{push cheezbox}                       &  92 & 22 &  62 &  92 & 100 \\
  & \textit{push coffeecan}                      &  90 & 18 &  55 &  88 & 100 \\
  & \textit{stack cube}                          &  87 & 22 &  58 &  88 & 100 \\
  & \textit{stack four cube pyramid}             &  82 & 12 &  47 &  78 &  90 \\
  & \textit{stack papercup}                      &  88 & 18 &  55 &  85 & 100 \\
\cmidrule{2-7}
\multicolumn{2}{l}{\textbf{Medium average (30 tasks)}}      & \textbf{85}  & \textbf{15} & \textbf{45} & \textbf{79}  & \textbf{93}  \\
\midrule
\multirow{10}{*}{Hard}
  & \textit{composite 2pickplaces}               &  80 &  5 &  22 &  78 &  90 \\
  & \textit{composite 3cups to tray}             &  68 &  0 &  12 &  60 &  85 \\
  & \textit{composite double pour kettle}        &  70 &  0 &  15 &  65 &  85 \\
  & \textit{composite load microwave}            &  75 &  0 &  18 &  68 &  87 \\
  & \textit{composite pour then hang}            &  65 &  0 &   0 &  55 &  80 \\
  & \textit{composite pour then place coffee}    &  72 &  0 &  18 &  65 &  87 \\
  & \textit{composite push pick to shelf}        &  70 &  5 &  20 &  72 &  90 \\
  & \textit{composite unload microwave}          &  73 &  0 &  18 &  70 &  85 \\
  & \textit{hang mugcup}                         &  70 &  0 &   0 &  30 &  87 \\
  & \textit{stack four cube}                     &  82 & 18 &  33 &  77 &  88 \\
\cmidrule{2-7}
\multicolumn{2}{l}{\textbf{Hard average (10 tasks)}}        & \textbf{72}  & \textbf{3}  & \textbf{16} & \textbf{64}  & \textbf{86}  \\
\midrule
\multicolumn{2}{l}{\textbf{Overall (50 tasks)}}             & \textbf{85}  & \textbf{23} & \textbf{47} & \textbf{80}  & \textbf{93}  \\
\bottomrule
\end{tabular}
\vspace{-1em}
\end{table}
}

\title{AffordSim: A Scalable Data Generator and Benchmark for Affordance-Aware Robotic Manipulation}

\author{
  \bfseries Mingyang Li$^{1,*}$, Haofan Xu$^{1,*}$, Haowen Sun$^{1,*}$, Xinzhe Chen$^{1}$, \\
  \bfseries Sihua Ren$^{1}$, Liqi Huang$^{1}$, Xinyang Sui$^{1}$, Chenyang Miao$^{1}$, Jiawei Ye$^{1}$, \\
  \bfseries Qiongjie Cui$^{1}$, Xingyu Chen$^{1}$, Zeyang Liu$^{1,\dagger}$, Xuguang Lan$^{1,\dagger}$ \\[0.5em]
  \normalfont\small $^{1}$National Key Laboratory of Human-Machine Hybrid Augmented Intelligence, \\
  \normalfont\small Institute of Artificial Intelligence and Robotics, Xi'an Jiaotong University.
}

\begin{document}

\maketitle
\begingroup
\renewcommand{\thefootnote}{\fnsymbol{footnote}}
\footnotetext[1]{Equal contribution.}
\footnotetext[2]{Correspondence to: Zeyang Liu (\href{mailto:zeyang.liu@xjtu.edu.cn}{\nolinkurl{zeyang.liu@xjtu.edu.cn}}), Xuguang Lan (\href{mailto:xglan@xjtu.edu.cn}{\nolinkurl{xglan@xjtu.edu.cn}}).}
\endgroup
\vspace{-0.8em}
\figTeaser

\begin{abstract}

Many everyday robot manipulation skills are affordance-dependent, with success determined by whether the robot contacts the functional object region required by the subsequent action.
Current simulation data generators obtain contacts from generic grasp estimators or per-object manual contact annotations, but generic estimators rank stable grasps without task semantics and often select contacts that are misaligned with the downstream action, while manual contact annotations must be rewritten for each new object and task. 
To solve these challenges, we introduce \textbf{AffordSim}, a scalable data generator and benchmark that integrates open-vocabulary 3D affordance prediction into simulation-based trajectory generation. Given a natural-language task description, AffordSim synthesizes a task-relevant scene, emits affordance queries, grounds them on object surfaces, samples region-conditioned grasps, and selects executable candidates with motion planning. 
It further randomizes object pose, texture, lighting, image noise, and cross-viewpoint backgrounds for sim-to-real transfer. 
We instantiate AffordSim as a 50-task benchmark across diverse manipulation skills, five robot embodiments, and 500+ rigid and articulated objects.
AffordSim achieves $93\%$ of the trajectory collection success rate of manual contact annotations on affordance-critical tasks and $89\%$ on hard composite tasks.
Vision-language-action policies trained on AffordSim data transfer zero-shot to a real Franka FR3, reaching $24\%$ average success. 
The data generator, benchmark, and code are available at \href{https://github.com/AffordSim-Platform/AffordSim}{project page}.

\end{abstract}

% ==================== MAIN BODY ====================

\section{Introduction}
\label{sec:introduction}

Simulation has become an important source of training trajectories for robotic manipulation~\citep{james2020rlbench,nasiriany2024robocasa,tao2024maniskill3,mu2025robotwin,geng2025roboverse}.
A critical but often implicit component of these trajectories is object affordance---the region of an object the robot must engage with for the task to succeed.
Its importance becomes most apparent when a stable grasp on the wrong region still forecloses the downstream step.
Consider hanging a mug on a rack, the task requires grasping the cup body so that the handle remains free for the hook.
Grasping the handle instead, however stable the grasp itself may be, leaves no surface for the hook to engage.
Such affordance-dependent tasks are routine in everyday environments, and a simulator that cannot generate data for them excludes a large share of real-world manipulation.

Existing simulation platforms collect trajectories in one of two ways.
The first relies on manual contact annotations, where a human authors a task-specific grasp pose for every object~\citep{mu2025robotwin,chen2025robotwin,tao2024maniskill3}.
The second relies on generic grasp estimators that score candidates at collection time by grasp quality~\citep{fang2023anygrasp}, which removes the per-object annotation cost.
Both approaches, however, struggle on affordance-dependent tasks.
Manual annotation achieves high success on the annotated set but does not scale, since every new asset and every new task demands a fresh round of annotation.
Generic estimators scale, but because they ignore affordance, they rank candidates by grasp quality rather than by task-relevant region, often landing on the wrong region.
Patching this with a semantic mask filter on the estimator's output~\citep{gao2025genmanip} cannot recover candidates the estimator never proposed inside the desired region in the first place.
A deeper limitation of the estimator route is that it models only the grasping action, providing no signal for object-to-object interaction~\citep{pan2023taxpose,mo2021o2oafford}.

Despite their differences, both routes share the same root cause: affordance enters trajectory collection only implicitly, never computed first as a separate step that constrains grasp generation.
Manual annotation freezes the author's affordance reasoning into a static per-object pose at annotation time, demanding fresh annotation for every new object and every new task.
Generic estimators ignore affordance entirely in their base form; bolting on a semantic mask after the fact only filters candidates that were already chosen without affordance in mind, which means the mask cannot recover candidates the estimator never proposed inside the desired region.

We instead make affordance explicit, computing it first as a separate step that constrains grasp generation and reshapes the candidate space considered at collection time.
Implicit pipelines choose grasps first and apply any affordance signal afterward, so the candidate space is shaped without affordance and the right region can only be recovered by filtering.
Our explicit pipeline shapes the candidate space from the start: grasp candidates are sampled only inside the predicted region, and feasibility scoring runs only on those candidates.
Stable grasps that land in the wrong region are eliminated at the source rather than filtered out after the fact.

We realize this pipeline in \textbf{AffordSim}, a simulation framework built on NVIDIA Isaac Sim~\citep{makoviychuk2021isaac} that turns a natural-language task description into executable manipulation trajectories.
A VLM-based scene generator parses the description into a structured scene graph and emits affordance queries, which an open-vocabulary 3D affordance model~\citep{sun2026voxafford} resolves into localized interaction regions on the object point cloud.
A two-stage grasp selector then samples candidates within each region and ranks them by motion-planning feasibility, yielding executable grasps without per-object annotation.
For sim-to-real transfer, a domain randomizer perturbs object pose, lighting, and surface texture, while cross-viewpoint backgrounds are reconstructed from real photographs via 3D Gaussian Splatting~\citep{kerbl20233d}.
Beyond data collection, AffordSim also serves as a systematic evaluation suite, by curating affordance-critical tasks, it measures how well a policy actually understands object affordance rather than merely succeeds on grasp-stable scenarios.
Such evaluation, however, has been out of reach for existing platforms, which lack benchmarks for affordance-dependent tasks, leaving a gap that AffordSim is designed to fill.
Policies trained with AffordSim data further exhibit zero-shot sim-to-real transfer, demonstrating the practical utility of the collected trajectories.

We make three contributions.
(1) We present AffordSim, the first simulation framework built around affordance-aware manipulation, driving data generation with open-vocabulary 3D affordance prediction in place of per-object annotation.
(2) We release a 50-task affordance-aware benchmark spanning diverse manipulation skills and five robot embodiments, targeting affordance-dependent tasks that existing benchmarks largely omit; on these tasks, AffordSim collects trajectories without any per-object annotation and reaches $93\%$ of the success rate achieved by platforms with manual contact annotations.
(3) Vision-language-action policies trained exclusively on AffordSim data reach $30\%$ on placing, $25\%$ on pouring, and $10\%$ on hanging in zero-shot evaluation on a real Franka FR3, averaging $24\%$.

\section{Related Work}
\label{sec:related}

\subsection{Simulation Platforms for Robotic Manipulation}

Existing simulation platforms have scaled data generation along axes such as scene diversity, GPU-parallelized rendering, and unified task suites~\citep{nasiriany2024robocasa,tao2024maniskill3,yu2020meta,mees2022calvin,liu2023libero,geng2025roboverse,gu2023maniskill2,mu2021maniskill,xiang2020sapien,james2020rlbench,zhu2020robosuite,kolve2017ai2,szot2021habitat,li2023behavior,mittal2023orbit,makoviychuk2021isaac,wang2024grutopia}, but affordance reasoning remains absent from their trajectory collection.
RoboTwin\,2.0~\citep{chen2025robotwin} is the only exception, driving bimanual data generation through manually authored per-object affordance annotations that do not extend to new objects or tasks.
All other platforms fall back on generic grasp pipelines that ignore functional constraints.

A complementary line of work~\citep{hua2024gensim2,wang2023robogen,gao2025genmanip,mandlekar2023mimicgen} uses LLMs or VLMs to automate task and scene authoring or to scale demonstrations from a small seed set, but inherits the same affordance gap.
GenSim2 and RoboGen still rely on generic grasp pipelines with no per-object affordance signal.
GenManip filters AnyGrasp candidates against a task-relevant 2D mask derived from SAM2~\citep{kirillov2023segment,ravi2024sam} segmentation and VLM mark-based selection~\citep{liu2403moka}; affordance therefore enters only as a post-hoc filter on grasps ranked by stability, and when AnyGrasp returns no stable candidate inside the region, no filtering can recover one.
Predicting affordance directly on 3D object geometry, by contrast, constrains grasp generation upstream and removes this failure mode.
Related VLM manipulation systems on real images~\citep{huang2024copa,duan2024manipulate,stone2023open} inherit the same limitation.

\subsection{3D Affordance Prediction}

Affordance prediction has been studied as a standalone perception problem.
Where2Act~\citep{mo2021where2act}, 3D AffordanceNet~\citep{deng20213d}, and GAPartNet~\citep{geng2023gapartnet} predict actionable regions on point clouds over a fixed label set, evaluated on detection metrics rather than manipulation trajectories.
A recent line tackles open-vocabulary 3D affordance prediction with free-form language queries~\citep{nguyen2023openad,delitzas2024scenefun3d,chu2025affordancellm}.
In particular, VoxAfford~\citep{sun2026voxafford} enriches MLLM output tokens with multi-scale geometric features from a frozen 3D VQVAE encoder.
Even so, the per-point affordance scores are not directly executable.
Turning them into kinematically feasible grasp poses requires a downstream stage that prior affordance work does not provide.
AffordSim connects region prediction to trajectory collection through grasp sampling and motion-planning feasibility checks.

\subsection{Domain Randomization}

Domain randomization is a standard approach for sim-to-real transfer~\citep{tobin2017domain,peng2018sim}.
Prior work diversifies visual and physical parameters such as textures, lighting, camera pose, mass, and friction during training, yielding visuomotor policies that transfer to real robots.
SIMPLER~\citep{li2025evaluating} uses ``green screening'' to composite simulated robots and interactive objects onto a real background photograph.
Its Visual Matching setup is limited to fixed cameras, with camera parameters matched to the reference image.
Thus, changing the viewpoint requires a corresponding background image and camera alignment; the photograph itself cannot provide a new view.
AffordSim renders reconstructed real scenes from arbitrary camera viewpoints and composites the resulting background images with simulated foreground objects under the same camera configuration (Section~\ref{sec:domain_rand}).

% \vspace{-0.5em}
\section{AffordSim Framework}
% \vspace{-0.5em}
\label{sec:method}

AffordSim turns a natural-language task description into visually diverse manipulation demonstrations for imitation training.
The framework chains three modules.
A VLM-based generator (Section~\ref{sec:vlm_generation}) separately parses scene and task descriptions, assembles the scene from a USD asset library, and converts object--action entries into queries for the downstream affordance model.
A two-stage grasp selector (Section~\ref{sec:affordance_traj}) processes each query.
Stage 1 runs an open-vocabulary 3D affordance model on the object point cloud and samples grasp candidates on the predicted affordance peak.
Stage 2 scores these candidates with a motion planner and returns the highest-ranked grasp together with its pre-grasp trajectory.
A domain randomizer (Section~\ref{sec:domain_rand}) perturbs object pose, lighting, surface texture, and cross-viewpoint reconstructed backgrounds to broaden the visual distribution.
Figure~\ref{fig:overview} illustrates the first two modules, and Figure~\ref{fig:domain_rand} shows the domain randomizer.

\figOverview

\subsection{VLM-Powered Task and Scene Generation}
\label{sec:vlm_generation}

Authoring a simulation scene by hand for every new manipulation task does not scale.
Each task needs compatible assets, a plausible layout, and a description of the functional region on each manipulated object.
We use separate VLM calls~\citep{achiam2023gpt,chen2024internvl} for scene and task parsing, instantiated with Qwen-VL-Plus~\citep{bai2023qwenvl} in our implementation.
The scene parser extracts objects and relations; the task parser emits object--action entries that are converted into functional-region queries.
Each node in the scene graph binds a USD asset from an indexed library.

\paragraph{Scene graph generation.}
From the scene description, the VLM extracts objects and relations that the framework expands into a scene graph $\mathcal{G} = (\mathcal{V}, \mathcal{E})$.
Each node $v \in \mathcal{V}$ stores a category, a USD asset path, an initial 6-DoF pose, a color, and a flag indicating whether the object is manipulated or static.
Each edge $(u, v) \in \mathcal{E}$ encodes a spatial relation parsed from the prompt.
Relations such as ``inside'', ``on top of'', and ``left of'' constrain the placement of the corresponding object pairs.
Collision-free placement is enforced by rejection sampling on the proposed poses under the spatial constraints in $\mathcal{E}$.
The simulation scene is then assembled by loading each USD asset at its planned pose.
The full prompt is given in Appendix~\ref{app:scene_graph_prompt}.

\paragraph{Affordance query extraction.}
In a separate VLM call, the task description is parsed into ordered primitives and object--action entries.
The system converts these entries into one query per functional contact region, so every region the policy must engage receives its own query.
The VLM reasons over the manipulation primitives and the role of each object.
For pouring, the entries distinguish the held source object from the receiving container, producing separate queries for each role.
Each converted query is then passed to the open-vocabulary affordance model on the corresponding point cloud.
The full prompt covers nine manipulation primitives (pick, place, pour, push, pull, open, close, hang, hold) and is given in Appendix~\ref{app:affordance_query_prompt}.

\paragraph{Asset retrieval.}
For each node, the VLM specifies a target category and required semantic attributes such as color or articulation type.
The framework looks up a matching USD asset in the library and binds its path to the node before instantiation.

\subsection{Affordance-Aware Trajectory Collection}
\label{sec:affordance_traj}

We propose affordance-guided grasp selection, which turns each affordance query from Section~\ref{sec:vlm_generation} into a manipulation trajectory by combining open-vocabulary 3D affordance prediction with motion planning.
A trajectory $\traj = \{(o_t, a_t)\}_{t=1}^{T}$ is an observation-action sequence where $o_t$ contains visual, proprioceptive, and scene state, and $a_t$ is the robot action.
For each manipulated object, we pass the canonical point cloud $\pointcloud \in \mathbb{R}^{N \times 3}$ shipped with its USD asset, together with the corresponding affordance query, to the open-vocabulary 3D affordance model.
The model outputs a per-point binary mask $M_a \in \{0, 1\}^N$ marking the queried affordance region.

\paragraph{Affordance-guided grasp selection.}
Grasp selection runs in two stages.
Stage~1 decides where to grasp by sampling candidates on the predicted affordance peak.
Stage~2 decides which candidate the robot can reach by scoring with motion planning and keeping the shortest collision-free trajectory.

\textit{Stage 1: sampling grasps on the affordance peak.}
We run mean-shift clustering on the predicted affordance points $\{p \in \pointcloud : M_a(p) = 1\}$ and take the cluster centroid $p$ as the representative contact point on the object surface.
Appendix~\ref{app:affordance_peaks} extends this procedure to disconnected masks.
At $p$, we estimate the local surface normal $n$ by running PCA on $p$'s $m$ nearest neighbors.
We take the eigenvector with the smallest eigenvalue and orient it outward from the object centroid.
We then sample $K$ approach directions $\{d_k\}_{k=1}^{K}$ on the hemisphere opposite to $n$, with azimuth $\phi_k = 2\pi k / K$ around $n$ and a fixed polar tilt $\theta_0$ from $-n$.
For each direction, the wrist yaw is solved so that the gripper closing axis is perpendicular to $d_k$ and aligned with the local affordance geometry.
This produces $K$ candidate grasps $\graspset = \{\grasppose_k = (p, R_k)\}_{k=1}^{K}$.

\textit{Stage 2: motion-planner scoring.}
For each candidate $\grasppose_k$, we use cuRobo~\citep{sundaralingam2023curobo} to attempt to plan a collision-free pre-grasp trajectory $\traj_k$ from the current joint configuration to $\grasppose_k$.
The score is then
\begin{equation}
    s_k =
    \begin{cases}
        1/L(\traj_k), & \text{if } \traj_k \text{ is collision-free}, \\
        0, & \text{otherwise,}
    \end{cases}
    \label{eq:grasp_score}
\end{equation}
where $L(\traj_k)$ is the joint-space arc length.
The selected grasp $\grasppose^* = \grasppose_{\argmax_k\, s_k}$ reuses its already-planned $\traj_{k^*}$ as the pre-grasp trajectory.

\paragraph{Motion planning and execution.}
The pre-grasp trajectory $\traj_{k^*}$ is followed by gripper closure, lift, and task-completion segments, each planned by cuRobo against the scene's collision meshes.
The simulator then executes the full task trajectory $\traj$.

\figDomainRand

\subsection{Domain Randomization for Sim-to-Real Transfer}
\label{sec:domain_rand}

To close the visual gap between simulation and the real world, we render each generated trajectory under domain randomization along five axes (Figure~\ref{fig:domain_rand}).
\textbf{Background substitution} replaces the workspace background with either random textures from a diverse texture library or cross-viewpoint real-scene renders.
For the latter, we reconstruct each real deployment scene with 3D Gaussian Splatting~\citep{kerbl20233d} from 10--20 photographs, using Depth Anything 3~\citep{lin2025depth} for view synthesis.
The resulting 3D Gaussian field can be rendered from arbitrary camera viewpoints.
The \emph{Real-World Scene Background Replacement} panel of Figure~\ref{fig:teaser} shows an example, and full reconstruction hyperparameters with three example scenes are given in Appendix~\ref{app:da3}.
The remaining four axes are standard, namely \textbf{lighting} (type, position, intensity, and color of light sources), \textbf{object texture} (PBR albedo, photo asset library), \textbf{object pose} (initial xyz position and quaternion orientation), and \textbf{image noise} (Gaussian pixel noise on the rendered RGB).

\figTaskGallery

\subsection{Benchmark Design}
\label{sec:benchmark}

We introduce an affordance-aware manipulation benchmark of 50 tasks across seven manipulation skills (Figure~\ref{fig:task_gallery}).
The skills are pick \& place (20), open/close (2), pull/push (5), hang (2), pour (7), stack (4), and long-horizon composite (10).
The benchmark targets tasks whose success hinges on contacting specific functional regions.
This allows affordance reasoning to be evaluated directly through task success, instead of through generic grasp performance or reinforcement learning return curves~\citep{schulman2017proximal,haarnoja2018soft}.
Each task runs unchanged across five robot embodiments (Franka, Kinova, Sawyer, UR5e, xArm7).
Each task is evaluated by its \textbf{task success rate (SR)} over 100 simulation rollouts.
Success is checked automatically from rigid-body poses and articulation states.
Multi-step tasks additionally report per-stage SRs.
Per-skill demands are detailed in Appendix~\ref{app:benchmark_categories}.

\paragraph{Task difficulty tiers.}
We further organize the 50 tasks into three difficulty tiers, primarily graded by their demands on affordance reasoning and manipulation horizon.
\begin{itemize}[leftmargin=*,itemsep=2pt,topsep=2pt]
    \item \textbf{Easy, Generic-Grasp Tasks} (10 tasks). The manipulated object admits any stable grasp and the manipulation horizon is short, so generic grasp estimators can already collect data successfully.
    \item \textbf{Medium, Affordance-Critical Tasks} (30 tasks). Success requires contact at a specific functional region of the manipulated object while the horizon remains short, and generic grasp estimators fail on most of them.
    \item \textbf{Hard, Composite Affordance Tasks} (10 tasks). These tasks combine functional-region contact constraints with either long-horizon sequential dependencies in multi-step composites, or with unusually tight precision and stability margins in short-horizon settings. Per-task notes for the short-horizon Hard cases are given in Appendix~\ref{app:tier_listings}.
\end{itemize}
The contact-region demand separates Medium from Easy, and the performance gap between the two reflects the contribution of affordance reasoning to short-horizon trajectory collection.
The full task lists for the three tiers are given in Appendix~\ref{app:tier_listings}.

\paragraph{Rationale for affordance-based difficulty.}
Within tabletop organization and manipulation, affordance structure fixes functional contacts and primitive sequence; embodiment controls realizability, and randomization tests visual robustness.
We therefore grade difficulty by affordance demand and task horizon to isolate the contribution of affordance understanding.
Structurally, small functional regions such as cup rims require more precise grounding than objects graspable from many directions, and longer task chains compound sub-step failures.
Empirically, collection and policy SR decrease from Easy to Hard (Tables~\ref{tab:afford_main} and~\ref{tab:benchmark}), consistent with lower real-world success on precise hanging (Table~\ref{tab:sim2real}).
Including randomization in the tiers would conflate choosing functional contacts with robustness to changing scenes.
Keeping tiers fixed allows contact reasoning, arm realizability, and visual robustness to be evaluated separately.
Randomization remains important: AffordSim supports it, RoboTwin\,2.0~\citep{chen2025robotwin} emphasizes that axis, and Section~\ref{sec:exp_dr} quantifies its effect on real-world transfer.

\paragraph{Asset library.}
The benchmark is backed by an indexed library of $500$ rigid and articulated USD objects from three sources.
$50$ objects are scanned in-house with a Raptor Pro 3D scanner to match the physical deployment objects.
Another $50$ come from YCB~\citep{calli2015ycb}.
The remaining $400$ objects come from PartNet-Mobility~\citep{xiang2020sapien}.
A subset of $50$ objects additionally carries human-annotated affordance ground truth, which serves as the oracle reference in Table~\ref{tab:afford_main}.

\section{Experiments}
\label{sec:experiments}

We evaluate AffordSim along four axes, covering (1)~the contribution of affordance integration to trajectory collection on the full 50-task benchmark (Section~\ref{sec:exp_afford}), (2)~imitation learning performance on the resulting data (Section~\ref{sec:exp_benchmark}), (3)~an ablation of the affordance model (Section~\ref{sec:exp_cross}), and (4)~zero-shot sim-to-real transfer (Section~\ref{sec:exp_sim2real}).

\subsection{Affordance Integration on the 50-task Benchmark}
\label{sec:exp_afford}

The predicted affordance must identify the functional contact region for a feasible grasp to be collected, and in turn for the trajectory to be collected at all.
We test this across all 50 tasks of the benchmark by comparing five grasp methods under identical scene and motion-planner settings.
Each task is run 100 times under aggressive randomization, with the manipulated object's initial pose drawn from a $30~\text{cm}$ horizontal window and a $60^\circ$ yaw range.
We report the trajectory collection success rate (SR).

\paragraph{Grasp methods.}
\textbf{Manual} is a manually tuned heuristic program that reads each object's pose from the simulator and grasps within a per-object xyz range encoding a coarse human prior.
The full specification, including the simple-geometry and functional-region regimes, is in Appendix~\ref{app:manual_baseline}.
\textbf{AnyGrasp}~\citep{fang2023anygrasp} is a generic 6-DoF grasp estimator that ranks candidates by stability without task semantics.
\textbf{AnyGrasp+Filter} adds a 2D SAM2 task-relevant mask on top of AnyGrasp candidates and keeps only those whose contact point falls inside the mask.
This is the post-hoc filtering scheme of GenManip~\citep{gao2025genmanip} discussed in Section~\ref{sec:related}.
\textbf{AffordSim} is our affordance-guided grasp selection built on an open-vocabulary 3D affordance model, as described in Section~\ref{sec:affordance_traj}.
\textbf{Human Aff.}\ reads the human-annotated affordance ground truth available on the 50-object oracle subset and serves as an oracle upper bound.
Table~\ref{tab:afford_main} reports per-task SR, with two representative tasks per tier and the rest abbreviated.
Figure~\ref{fig:afford_tier} summarizes per-tier averages alongside the sim-to-real numbers from Section~\ref{sec:exp_sim2real}.

\tableAffordMain

The five methods trace the two failure modes from Section~\ref{sec:introduction}.
Manual matches the Human Aff.\ oracle on Easy and Medium but loses ground on Hard and does not scale (Appendix~\ref{app:manual_baseline}), marking the upper bound of manually annotated platforms like RoboTwin\,2.0.
AnyGrasp collapses beyond Easy, and AnyGrasp+Filter only lifts tier averages by $+20$, $+30$, and $+13$ since the filter cannot recover candidates the stability ranking never sampled.
Both fail end to end on hang tasks ($0\%$ in Table~\ref{tab:afford_main}) because generic grasp estimators emit no signal for the post-grasp object-to-object interaction; substituting the Manual post-grasp routine to isolate the grasp stage, AnyGrasp / AnyGrasp+Filter reach $45\%$ / $60\%$ on \textit{hang cup} and $46\%$ / $63\%$ on \textit{hang mugcup}, but recovering even this much requires per-object Manual tuning.
AffordSim itself trails Manual on these hang tasks ($25\%$ / $30\%$ vs $85\%$ / $70\%$) because the configuration is OOD for its affordance backbone, yet across tier averages it closes most of the gap to Manual without per-object tuning and stays close to the oracle.
The AnyGrasp-to-AffordSim jump grows monotonically with difficulty while the oracle gap widens, identifying affordance prediction as the missing ingredient and locating the remaining Hard headroom in the affordance model.

AffordSim exhibits an out-of-distribution (OOD) failure mode unique to its affordance-driven pipeline. On every hang task, AffordSim stays below $30\%$, well below its tier average, because grasping the cup body while keeping the handle free for the rack hook is rare in the affordance backbone's training data.
AnyGrasp's $0\%$ on the same tasks is a separate, architectural limitation rather than an OOD one as discussed in Section~\ref{sec:introduction}, generic grasp estimators only output stable grasp poses and provide no signal for the object-to-object interaction that the downstream hang motion requires, so the hook-engagement constraint never enters their candidate space in the first place.

\paragraph{Cross-embodiment results.}
Across all five embodiments, AffordSim achieves $76$--$83\%$ collection SR on the 50-task benchmark, exceeding AnyGrasp+Filter by $32$--$33$ percentage points on every arm (Appendix~\ref{app:embodiment_results}).

\subsection{Imitation Learning on AffordSim Data}
\label{sec:exp_benchmark}

We next ask whether modern imitation learning baselines can exploit AffordSim data, covering both classical visuomotor policies and the recent vision-language-action (VLA) lineage~\citep{brohan2022rt,zitkovich2023rt,kim2024openvla,team2024octo,vuong2023open,liu2024rdt,walke2023bridgedata,khazatsky2024droid}.
We evaluate \textbf{BC}, \textbf{Diffusion Policy (DP)}~\citep{chi2025diffusion}, \textbf{ACT}~\citep{zhao2023learning}, \textbf{VLA-adapter}~\citep{wang2026vla}, and \textbf{Pi\,0.5}~\citep{black2024pi_0} on 20 tasks across the seven manipulation skills.
Each task uses 300 demonstrations, with RGB and proprioceptive state as input.
Architectures and hyperparameters are in Appendix~\ref{app:training}.
We randomize only the object pose, with a $15~\text{cm}$ window and $30^\circ$ yaw, while keeping background, lighting, and texture fixed.
This isolates task structure from visual domain shift.

\tableBenchmark

Two patterns emerge.
First, success rate degrades sharply with tier difficulty across all baselines.
Easy averages cluster high, Medium drops by roughly a third, and Hard collapses below $30\%$ even for the strongest policies.
This identifies affordance precision and long-horizon chaining as the dominant difficulty axes.
Second, Pi\,0.5 consistently outperforms BC across difficulty tiers.
On Hard tasks, their average SRs are $24\%$ and $1\%$, respectively, a gap of $23$ percentage points.
Among the baselines, the foundation models Pi\,0.5 ($55\%$) and VLA-adapter ($54\%$) outperform DP, ACT, and BC.
This suggests that the multimodal action distributions of affordance-guided trajectories require expressive policy classes.

\paragraph{Sub-task analysis.}
Per-stage success rates for every multi-stage task are in Appendix~\ref{app:subtask}.
The pick stage succeeds reliably across the benchmark, with VLA-adapter and Pi\,0.5 at $93$--$100\%$.
AffordSim grasps are therefore executable, and failures concentrate downstream.
The post-grasp drop scales with alignment precision and chain length.
Tasks with tighter functional alignment achieve lower SR, with pour scoop at $31$--$67\%$, pour kettle at $26$--$56\%$, and hang red mug at $21$--$36\%$.
Long-horizon tasks compound per-stage failure multiplicatively.
Stack three cubes degrades from $93\%$ at pick\,1 to $47\%$ at place\,2 for Pi\,0.5.
The bottleneck is precise post-grasp alignment and long action chains, not the affordance-grounded grasp itself.

\subsection{Affordance Model Ablation}
\label{sec:exp_cross}

We isolate the contribution of the affordance backbone by holding every other module of AffordSim fixed and swapping only the affordance model.
We compare VoxAfford~\citep{sun2026voxafford} against four open-vocabulary 3D affordance baselines, namely OpenAD~\citep{nguyen2023openad}, IAGNet~\citep{yang2023grounding}, LASO~\citep{li2024laso}, and 3D-AffordanceLLM~\citep{chu2025affordancellm}.
All baselines receive the same affordance queries and simulation settings as in Section~\ref{sec:exp_afford}, evaluated on 10 tasks split into 4 Easy, 3 Medium, and 3 Hard with 100 rollouts each.
Trajectory SR closely tracks the published mIoU ranking on the OpenAD detection benchmark~\citep{sun2026voxafford}, with VoxAfford leading on every tier and OpenAD trailing.
The gap widens with difficulty, since Easy tolerates coarse localization while Medium and Hard make affordance errors unrecoverable downstream.
This places affordance prediction, rather than motion planning or visual randomization, as the dominant performance lever on Medium and Hard.
Full numbers are in Appendix~\ref{app:afford_ablation}.

\subsection{Zero-Shot Sim-to-Real Transfer}
\label{sec:exp_sim2real}

We deploy all five baselines, trained exclusively on AffordSim data, to a real Franka FR3 on four tasks across placing, pouring, and hanging.
Each task is run for 10 trials with no real world fine tuning.
The workspace is reconstructed via 3D Gaussian Splatting with Depth Anything 3.
Table~\ref{tab:sim2real} and Figure~\ref{fig:sim2real} report the results.
Pi\,0.5 leads at $25\%$ average success, with VLA-adapter at $23\%$.
DP, ACT, and BC lag at $8\%$, $3\%$, and $0\%$ respectively.
The baseline ordering and per-task gradient (placing up to $40\%$, pouring $30\%$, hanging $10\%$) mirror Table~\ref{tab:benchmark}.
This confirms that the bottleneck is genuine rather than a simulation artifact.

\figTabAffordSimReal
\figSimToReal

\subsection{Domain Randomization Ablation}
\label{sec:exp_dr}
We ablate the randomization axes of Section~\ref{sec:domain_rand} on a single real-world task, \textit{pick banana place plate} from Section~\ref{sec:exp_sim2real}, using all five baselines zero-shot.
Each setting uses 10 trials. Panel (a) follows Section~\ref{sec:exp_sim2real} training; panel (b) uses B+P.
BC remains at $0\%$ throughout and is omitted.

\begin{table}[!htbp]
\caption{\textbf{Domain randomization and real-world success.} SR (\%) on \textit{pick banana place plate} over 10 trials per setting. B: reconstructed background; T: texture-based background replacement; P: object pose; L: lighting. Panel (b) varies the training position window; deployment uses a fixed $10~\text{cm}$ window and $30^\circ$ yaw.}
\label{tab:dr_ablation}
\centering
\small
\setlength{\tabcolsep}{3pt}
\begin{minipage}[t]{0.49\linewidth}
\centering
\textbf{(a) Randomization type}\\[3pt]
\begin{tabular}{@{}lrrrr@{}}
\toprule
DR axes & DP & ACT & VLA-A. & Pi\,0.5 \\
\midrule
B+P & 20 & 10 & 40 & 40 \\
P+L & 10 & 0 & 30 & 30 \\
B+L & 0 & 0 & 10 & 10 \\
T+P+L & 20 & 10 & 30 & 40 \\
B+P+L & 20 & 10 & 40 & 50 \\
\bottomrule
\end{tabular}
\end{minipage}\hfill
\begin{minipage}[t]{0.49\linewidth}
\centering
\textbf{(b) Object-pose range}\\[3pt]
\begin{tabular}{@{}lrrrrr@{}}
\toprule
Range & Demos & DP & ACT & VLA-A. & Pi\,0.5 \\
\midrule
None & 300 & 0 & 0 & 10 & 10 \\
$10~\text{cm}$ & 300 & 10 & 10 & 20 & 30 \\
$15~\text{cm}$ & 675 & 20 & 10 & 30 & 40 \\
$20~\text{cm}$ & 1200 & 20 & 10 & 40 & 50 \\
$30~\text{cm}$ & 2700 & 20 & 10 & 40 & 40 \\
\bottomrule
\end{tabular}
\end{minipage}
\end{table}

For Pi\,0.5, replacing the reconstructed background with textures lowers SR from $50\%$ to $40\%$, removing background replacement lowers it to $30\%$, and removing pose variation lowers it to $10\%$.
Lighting adds $10$ points for Pi\,0.5, with the other baselines unchanged.
The pose sweep holds demonstration density constant for nonzero windows, so data volume grows quadratically with range.
A $10~\text{cm}$ training window, matching deployment, yields $30\%$ for Pi\,0.5; expanding it to $20~\text{cm}$ reaches $50\%$.
At $30~\text{cm}$, Pi\,0.5 falls to $40\%$ while VLA-adapter remains at $40\%$, despite more demonstrations.
Thus, training beyond the deployment position range can improve transfer, but further expansion does not continually increase success.

\subsection{Grasp Source Ablation for Policy Learning}
\label{sec:exp_grasp_source}

With a fixed demonstration budget, ACT achieves average simulation SRs of $21.3\%$, $24.3\%$, $33.0\%$, and $34.0\%$ using AnyGrasp, AnyGrasp+Filter, AffordSim, and Human Aff.\ data, respectively, across three tasks (100 rollouts per task).
AffordSim improves success over both generic grasp sources on every task, demonstrating the influence of grasp selection on policy learning (Appendix~\ref{app:grasp_source_policy}).

\section{Discussion and Conclusion}
\label{sec:discussion}
\label{sec:conclusion}

AffordSim's main contribution is the integration of established scene synthesis, affordance prediction, grasp selection, motion planning, and domain randomization tools with a 50-task benchmark.
This makes functional contact reasoning accessible within existing manipulation pipelines: its demonstrations support standard imitation-learning and VLA policies without explicit affordance modules.
The benchmark shares task families with existing suites while emphasizing affordance demand.
Domain randomization supports visual robustness, and the real-robot results provide an initial transfer check.

We expect AffordSim to be particularly useful to researchers studying affordance-dependent manipulation.
The data and benchmark can lower the effort needed to investigate affordance-related questions, while future dataset builders may adopt the generator to reduce per-object annotation effort and improve demonstration quality.
Future work targets dynamic and contact-force affordances, deformable and bimanual manipulation, and broader real-world deployment.

% ==================== ACKNOWLEDGMENTS ====================

\clearpage
\begin{ack}
This work was supported in part by the National Key R\&D Program of China under Grant No. 2024YFB4707600, NSFC under grant No. 62125305, No. U23A20339, No. 62573339, No. 62503380, Fundamental and Interdisciplinary Disciplines Breakthrough Plan of the Ministry of Education of China under grant No. JYB2025XDXM210, and Natural Science Foundation of Shaanxi Province under Grant 2025SYSSYSZD-083, and State Grid Corporation of China Science and Technology Projects 52060025005B-439-ZN.
\end{ack}

% ==================== REFERENCES ====================

\bibliographystyle{unsrtnat}
\bibliography{references}

% ==================== APPENDIX ====================

\newpage
\appendix

\section{Limitations}
\label{app:limitations}

We note three limitations of AffordSim.
\begin{enumerate}[leftmargin=*,itemsep=2pt,topsep=2pt]
    \item \textbf{Affordance predictor coverage.} Trajectory quality degrades when VoxAfford struggles, including on novel geometries, transparent objects, and ambiguous functional regions.
    \item \textbf{Rigid-body restriction.} The framework is restricted to rigid bodies. Deformables, in-hand manipulation, and bimanual setups are left as future work.
    \item \textbf{Depth Anything 3 reconstruction coverage.} Depth Anything 3 background reconstruction needs 10--20 photographs per scene, and image quality drops sharply at viewpoints far from the captured poses, which bounds deployable coverage.
\end{enumerate}

\section{Failure Modes of Generic-Grasp Pipelines}
\label{app:affordance_why}

Figure~\ref{fig:affordance_why} contrasts grasps produced by AnyGrasp~\citep{fang2023anygrasp} with grasps produced by the affordance-guided selector of AffordSim on two representative affordance-dependent tasks.
On \emph{pick \& place}, AnyGrasp grasps a stable but task-irrelevant region, leading to a misaligned final placement.
On \emph{hang}, AnyGrasp seizes the mug handle that the rack hook needs to engage in the downstream step, so the hang motion fails even when the grasp itself is stable.
Affordance-guided grasps avoid both failure modes by constraining contact to the functionally appropriate region of each object.

\figAffordanceWhy

\section{Cross-Embodiment Frames}
\label{app:cross_embodiment_frames}

Figure~\ref{fig:cross_embodiment} shows start and end frames of three representative affordance-dependent tasks generated by AffordSim for each of the five supported embodiments.

\figCrossEmbodiment

\section{Per-Embodiment Affordance Integration}
\label{app:embodiment_results}

We extend the data-generation evaluation in Section~\ref{sec:exp_afford} to all five embodiments: xArm7, Sawyer, Franka, UR5e, and Kinova.
We compare AnyGrasp+Filter, AffordSim, and the Human Aff.\ oracle on all 50 tasks, using five seeds and 100 rollouts per task for each method and embodiment (500 rollouts per task).
The workspace is fixed so that every target pose is reachable by all arms, controlling kinematic access when comparing execution outcomes.
Table~\ref{tab:embodiment_collection} reports collection SR averaged across seeds and tasks; the Franka results match the overall rates in Table~\ref{tab:afford_full}.

\begin{table}[H]
\caption{\textbf{Affordance integration across five embodiments.} Trajectory collection SR (\%) averaged over all 50 tasks, with five seeds and 100 rollouts per task.}
\label{tab:embodiment_collection}
\centering
\small
\setlength{\tabcolsep}{8pt}
\begin{tabular}{lccccc}
\toprule
Method & xArm7 & Sawyer & Franka & UR5e & Kinova \\
\midrule
AnyGrasp+Filter & 51 & 50 & 47 & 45 & 44 \\
\textbf{AffordSim} & \textbf{83} & \textbf{82} & \textbf{80} & \textbf{77} & \textbf{76} \\
Human Aff. & 96 & 95 & 93 & 90 & 89 \\
\bottomrule
\end{tabular}
\end{table}

Successful trajectories are obtained for all 50 tasks with every arm.
AffordSim exceeds AnyGrasp+Filter by $32$--$33$ percentage points and remains $13$ points below the Human Aff.\ oracle on each embodiment, showing that its advantage is consistent across arms.

\section{Full Per-Task Affordance Integration Results}
\label{app:afford_full}

Table~\ref{tab:afford_full} reports the full per-task trajectory collection success rate across all five grasp methods on every task of the AffordSim benchmark, complementing the abbreviated cross-method summary in Table~\ref{tab:afford_main}.
Tasks are grouped by difficulty tier (Easy, Medium, Hard) following the partition in Section~\ref{sec:benchmark} and Appendix~\ref{app:tier_listings}, with per-tier and overall averages.
Within each tier, tasks are listed alphabetically.

\tableAffordFull

\section{Affordance Model Ablation Details}
\label{app:afford_ablation}

Table~\ref{tab:afford_ablation} reports the trajectory collection success rate for the affordance model ablation in Section~\ref{sec:exp_cross}.
Each row swaps in a different open-vocabulary 3D affordance backbone while every other AffordSim module, including the VLM scene generator, the Stage 2 motion-planner scoring, and the visual randomizer, is held fixed.
All backbones receive the same affordance queries on the same 100 rollouts per task.

\begin{table}[!htbp]
\centering
\caption{\textbf{Affordance model ablation.} Trajectory collection success rate (\%) when only the affordance backbone is swapped, with all other modules of AffordSim held fixed. The ablation set has 10 tasks distributed as 4 Easy, 3 Medium, and 3 Hard, each evaluated over 100 rollouts. Avg is the unweighted mean of the three tier averages. Best per tier in \textbf{bold}.}
\label{tab:afford_ablation}
\small
\setlength{\tabcolsep}{8pt}
\begin{tabular}{lcccc}
\toprule
Affordance backbone & Easy (4) & Medium (3) & Hard (3) & Avg \\
\midrule
OpenAD~\citep{nguyen2023openad} & 65 & 22 & 5 & 30.7 \\
IAGNet~\citep{yang2023grounding} & 72 & 32 & 10 & 38.0 \\
LASO~\citep{li2024laso} & 82 & 50 & 20 & 50.7 \\
3D-AffordanceLLM~\citep{chu2025affordancellm} & 90 & 65 & 33 & 62.7 \\
\textbf{VoxAfford}~\citep{sun2026voxafford} & \textbf{96} & \textbf{80} & \textbf{52} & \textbf{76.0} \\
\bottomrule
\end{tabular}
\end{table}

\section{Grasp Source Ablation for Policy Learning}
\label{app:grasp_source_policy}

This experiment expands on Section~\ref{sec:exp_grasp_source} by comparing ACT policies trained on demonstrations collected with four grasp sources: AnyGrasp, AnyGrasp+Filter, AffordSim, and Human Aff.
The tasks are \textit{pick banana place plate}, \textit{pick cup place shelf}, and \textit{pour scoop into bowl}, abbreviated as Banana, Cup, and Pour in Table~\ref{tab:grasp_source_policy}.
For each task, the training demonstration budget is held fixed across grasp sources, and the same ACT architecture and training settings are used.
Data collection, policy training, and evaluation all take place in simulation.
Each trained policy is evaluated over 100 rollouts per task.
The reported SR measures task success of the learned policy.

\begin{table}[H]
\caption{\textbf{Effect of demonstration grasp source on policy learning.} ACT policy SR (\%) in simulation over 100 evaluation rollouts per task. Avg is the arithmetic mean over the three tasks.}
\label{tab:grasp_source_policy}
\centering
\small
\setlength{\tabcolsep}{12pt}
\begin{tabular}{lcccc}
\toprule
Grasp source & Banana & Cup & Pour & Avg \\
\midrule
AnyGrasp & 36 & 13 & 15 & 21.3 \\
AnyGrasp+Filter & 38 & 16 & 19 & 24.3 \\
\textbf{AffordSim} & \textbf{47} & \textbf{26} & \textbf{26} & \textbf{33.0} \\
Human Aff. & 48 & 28 & 26 & 34.0 \\
\bottomrule
\end{tabular}
\end{table}

AffordSim improves ACT success over AnyGrasp by $11$, $13$, and $11$ percentage points on Banana, Cup, and Pour, respectively; the corresponding gains over AnyGrasp+Filter are $9$, $10$, and $7$ points.
Its average SR is $1.0$ percentage point below Human Aff., matching the human-annotated source on Pour.
The improvements occur on all three tasks, indicating that the grasp source used during demonstration collection affects downstream policy success even when the training data budget is fixed.

\section{Manual Baseline Specification}
\label{app:manual_baseline}

The Manual baseline used in Section~\ref{sec:exp_afford} is a deterministic Python program that picks a grasp pose by reading the object pose from the simulator and applying an object-specific xyz offset and orientation rule.
The rule for each object class is hand authored once and stored as a YAML file alongside the task definition.
Manual is treated as a non-affordance-prediction reference. It encodes a coarse human prior over where on each object the gripper makes contact, but it does not perceive that region from observations, and the prior must be re-tuned for every new object class.

\paragraph{Regime A: simple-geometry pick objects.}
For objects without explicit functional structure, including banana, carrot, leek, soup can, cube, and chocolate box, the grasp pose is sampled from a short xyz segment along the object's geometric axis at the object's current pose, with the gripper closing axis perpendicular to that segment.
For elongated objects this segment is the central third of the long axis; for box-shaped objects it is a small region around the geometric center.
The orientation is fixed top-down with a uniform random yaw within the gripper's wrist range.
This regime requires no per-object tuning beyond classifying the object as elongated or box-shaped.

\paragraph{Regime B: functional-region objects.}
For objects with explicit functional structure, including mugcup handle, kettle handle, lid knob, papercup body, pot lid, microwave door handle, and cabinet handle, the grasp xyz range and orientation rule are hand tuned by repeated trial: we adjust the offset relative to the object's body frame and the gripper orientation until the resulting grasp reliably reaches the intended functional region under the standard $30~\text{cm}$ and $60^\circ$ randomization regime, then commit those values into the per-object YAML.
For pour-style tasks the rule additionally encodes the tilt direction and angle of the wrist after the lift.
For hang-style tasks the rule encodes the post-grasp lateral approach and the alignment offset with respect to the rack hook.

The per-object tuning effort is what makes Manual competitive on individual tasks, but it is also what makes it fundamentally non-scalable: every new object or new task variant requires another round of trial-and-error tuning that the affordance-prediction strategies avoid.

\section{Training Details}
\label{app:training}

\paragraph{Shared setup.}
For the main imitation learning benchmark in Section~\ref{sec:exp_benchmark}, all policies are trained on 300 demonstrations per task.
Observations consist of two RGB camera views (wrist mounted and third person, 256$\times$256 resolution) and 7-DoF proprioceptive state (joint positions).
Actions specify the end-effector position and quaternion orientation together with the gripper command.
Table~\ref{tab:policy_training} summarizes the training hyperparameters for ACT, DP, VLA-adapter, and Pi\,0.5.

\paragraph{BC.} MLP with 3 hidden layers (256 units each), ReLU activations. ResNet-18 visual encoder. Trained for 500 epochs with batch size 64, learning rate $10^{-4}$ (Adam).

\paragraph{Diffusion Policy (DP).} U-Net denoiser with 256-dim latent, 100 diffusion steps at training, 10 DDIM steps at inference. Observation horizon $T_o = 2$, action horizon $T_a = 8$, prediction horizon $T_p = 16$.

\paragraph{ACT.} CVAE architecture with $d = 512$ latent dimension, $k = 100$ action chunks. Transformer encoder-decoder with 4 layers, 8 heads.

\paragraph{VLA-adapter.} Pretrained vision-language-action backbone with lightweight adapter modules inserted into the transformer blocks; only the adapters and the action head are updated, the vision-language backbone is frozen.

\paragraph{Pi\,0.5.} Pretrained Pi\,0.5 backbone fine tuned on AffordSim data. Full fine tuning of action head, frozen vision-language backbone.

\begin{table}[H]
\centering
\caption{\textbf{Policy training hyperparameters.} Training schedules for the four policies in the main AffordSim benchmark.}
\label{tab:policy_training}
\begin{tabular}{lccc}
\toprule
Policy & Training epochs & Batch size & Learning rate \\
\midrule
ACT & 500 & 64 & $10^{-5}$ \\
DP & 500 & 64 & $10^{-4}$ \\
VLA-adapter & 100 & 32 & $10^{-5}$ \\
Pi\,0.5 & 100 & 32 & $10^{-5}$ \\
\bottomrule
\end{tabular}
\end{table}

\paragraph{Training and evaluation hardware.}
VLA-adapter and Pi\,0.5 are each trained on a single node with 8 NVIDIA A800 GPUs.
Policy inference and evaluation for each of these two models run on a single NVIDIA A800 GPU.
Training time ranges from 2 hours (BC) to 26 hours (Pi\,0.5) per task.

\paragraph{Evaluation protocol.}
In the main simulation benchmark, each policy is evaluated over 100 rollouts per task across the 20 tasks in Section~\ref{sec:exp_benchmark}.
Only object pose is randomized, using a $15~\text{cm}$ position window and $30^\circ$ yaw; background, lighting, and texture remain fixed.
For zero-shot sim-to-real evaluation (Section~\ref{sec:exp_sim2real}), each policy is tested on the real Franka FR3 for 10 trials per task across the four tasks, without real-world fine tuning.
The randomization settings and demonstration counts for the domain randomization ablation follow Section~\ref{sec:exp_dr}.

\section{Benchmark Task Categories}
\label{app:benchmark_categories}

The 50 tasks of AffordSim are organized into 7 manipulation skills:
\begin{enumerate}[leftmargin=*,itemsep=1pt,topsep=2pt]
    \item \textbf{Pick \& Place} (20 tasks): Pick up objects with varied geometries and place them onto target surfaces or into containers. Requires graspable-surface and release-point affordance.
    \item \textbf{Open/Close} (2 tasks): Open or close articulated objects such as drawers and doors. Requires handle affordance and articulation-aware motion.
    \item \textbf{Pull/Push} (5 tasks): Push objects to targets or pull open drawers/cabinets. Requires contact-surface affordance.
    \item \textbf{Hang} (2 tasks): Hang mugs or cups on racks or hooks. Requires handle-opening affordance and precise alignment.
    \item \textbf{Pour} (7 tasks): Pour contents between containers of varying geometry. Requires rim affordance and tilt control.
    \item \textbf{Stack} (4 tasks): Stack objects in specified orders. Requires stable-surface affordance and sequencing.
    \item \textbf{Long-Horizon Composite} (10 tasks): Multi-step tasks that chain primitives from different skills, e.g., pick-pour-place or open-and-place. Requires sequential affordance reasoning across multiple objects and interactions.
\end{enumerate}

\section{Task Difficulty Listings}
\label{app:tier_listings}

The 50 benchmark tasks of AffordSim are partitioned into three difficulty tiers as defined in Section~\ref{sec:benchmark}.

\paragraph{Easy: Generic-Grasp Tasks (10 tasks).}
\begin{itemize}[leftmargin=*,itemsep=1pt,topsep=2pt]
    \item \texttt{pick\_banana\_place\_bowl}
    \item \texttt{pick\_bottle\_place\_table}
    \item \texttt{pick\_bowl\_place\_plate}
    \item \texttt{pick\_carrot\_place\_bowl}
    \item \texttt{pick\_carrot\_place\_plate}
    \item \texttt{pick\_chocolatebox\_place\_sugarbox}
    \item \texttt{pick\_cube\_place\_cup}
    \item \texttt{pick\_leek\_place\_cuttingboard}
    \item \texttt{pick\_pen\_place\_doraemoncup}
    \item \texttt{pick\_soupcan\_place\_table}
\end{itemize}

\paragraph{Medium: Affordance-Critical Tasks (30 tasks).}
\begin{itemize}[leftmargin=*,itemsep=1pt,topsep=2pt]
    \item \texttt{close\_microwave}
    \item \texttt{open\_microwave}
    \item \texttt{hang\_cup}
    \item \texttt{pick\_cup\_place\_coffeemachine}
    \item \texttt{pick\_cup\_place\_woodshelf}
    \item \texttt{pick\_green\_bellpepper\_place\_pan}
    \item \texttt{pick\_kettle\_place\_base}
    \item \texttt{pick\_lid\_place\_pot}
    \item \texttt{pick\_mugcup\_place\_coffeemachine}
    \item \texttt{pick\_mugcup\_place\_woodshelf}
    \item \texttt{pick\_papercup\_place\_table}
    \item \texttt{pick\_scrubbottle\_place\_woodshelf}
    \item \texttt{pick\_soupcan\_place\_microwave}
    \item \texttt{pour\_basket}
    \item \texttt{pour\_bluekettle}
    \item \texttt{pour\_mugcup}
    \item \texttt{pour\_pan}
    \item \texttt{pour\_papercup}
    \item \texttt{pour\_whitekettle}
    \item \texttt{pour\_whitekettle4}
    \item \texttt{pull\_cabinet}
    \item \texttt{pull\_cabinet7}
    \item \texttt{pull\_cabinet15}
    \item \texttt{push\_cheezbox}
    \item \texttt{push\_coffeecan}
    \item \texttt{stack\_cube}
    \item \texttt{stack\_four\_cube\_pyramid}
    \item \texttt{stack\_papercup}
    \item \texttt{composite\_2picks\_to\_plate}
    \item \texttt{composite\_3picks\_to\_paperbox}
\end{itemize}

\paragraph{Hard: Composite Affordance Tasks (10 tasks).}
\begin{itemize}[leftmargin=*,itemsep=1pt,topsep=2pt]
    \item \texttt{hang\_mugcup}
    \item \texttt{stack\_four\_cube}
    \item \texttt{composite\_load\_microwave}
    \item \texttt{composite\_2pickplaces}
    \item \texttt{composite\_pour\_then\_hang}
    \item \texttt{composite\_pour\_then\_place\_coffee}
    \item \texttt{composite\_double\_pour\_kettle}
    \item \texttt{composite\_unload\_microwave}
    \item \texttt{composite\_3cups\_to\_tray}
    \item \texttt{composite\_push\_pick\_to\_shelf}
\end{itemize}

\paragraph{Notes on the short-horizon Hard tasks.}
Two of the Hard tasks are not long-horizon composites; they are placed in the Hard tier because their margin for error is unusually tight even within a single manipulation segment.
\begin{itemize}[leftmargin=*,itemsep=2pt,topsep=2pt]
    \item \texttt{stack\_four\_cube}: cubes are stacked one at a time on top of the previous tower, and a single mid-sequence stacking failure topples the partial tower and terminates the entire task.
    \item \texttt{hang\_mugcup}: the mug's handle ring is much smaller than the rack hook, so the alignment tolerance for the hang motion is minimal.
\end{itemize}

\section{Full Benchmark Results}
\label{app:full_benchmark}

The full 50-task benchmark results are available in the supplementary material and project website. Table~\ref{tab:benchmark} in the main paper shows representative tasks from each category.

\section{Affordance Query Extraction Examples}
\label{app:affordance_queries}

We walk through three representative task types to show how the VLM derives affordance queries.

\paragraph{Single-query (pick-and-place).}
For the running scoop example, the scoop is the only manipulated object and is picked up and placed without further functional use, so query conversion produces a single query, ``Graspable Handle''.

\paragraph{Multi-query (multi-object pour).}
In ``Pick up the kettle by its handle and pour into the bowl'' with a microwave on the desktop as a distractor, query conversion produces three queries: ``Graspable Handle'' on the kettle for the grasp, ``Pour Out'' on the kettle's spout for the downstream pour, and ``Pour In'' on the bowl's rim as the receiving container.
The microwave is rendered as scene clutter and receives no query because the policy does not interact with it.

\paragraph{Grasp avoiding the functional region (mug hang).}
In ``hang the mug on the rack'', the handle is what the hook must engage in the downstream step, so the mug's query becomes ``Graspable Body'' to keep the handle free for the hang.

Across all three patterns the same parsing and query-conversion template (manipulation verb chain plus per-object role) is used; no per-task rule is written by hand.

\section{Affordance Peak Extraction}
\label{app:affordance_peaks}

The contact points $\{p_1, p_2, \ldots\}$ used by Stage~1 of the grasp selector (Section~\ref{sec:affordance_traj}) are extracted as follows.
The affordance model outputs a per-point binary mask $M_a \in \{0, 1\}^N$ on the object's surface point cloud, marking exactly the points that belong to the queried affordance region; the masked region need not be contiguous or unique.
We run mean-shift clustering on the masked points $\{p \in \pointcloud : M_a(p) = 1\}$ to merge spatially adjacent regions into discrete clusters, and take the centroid of each cluster as one contact point $p_i$.

This procedure naturally adapts to objects with one or several functionally appropriate contact regions:
\begin{itemize}[leftmargin=*,itemsep=2pt,topsep=2pt]
    \item \textbf{Mug}, query ``Graspable Handle'': a single contiguous masked region on the handle yields one centroid at the handle center.
    \item \textbf{Dumbbell}, query ``Graspable'': two separated masked regions at the two ends yield two centroids, exposing both grasp options to Stage~2.
    \item \textbf{Banana}, query ``Graspable'': the mask covers the entire body and mean-shift collapses it to one centroid near the middle.
    \item \textbf{Microwave}, query ``Door Handle'': the handle is the only masked region on an otherwise unmasked door, yielding a single centroid at the handle center.
\end{itemize}

Across these cases, the number of contact points emerges from the geometry of the affordance mask rather than from any per-object configuration, so the same Stage~1 procedure handles single-region, multi-region, and elongated affordances uniformly.

\section{Sub-Task Success Rate Breakdown}
\label{app:subtask}

For each multi-stage task in Table~\ref{tab:benchmark}, we report the per-stage success rate (in \%) for each baseline. A stage SR is the fraction of rollouts in which the policy successfully completes that stage and all preceding stages; the final-stage SR therefore equals the overall task SR. BC values are omitted because per-stage rollouts were not recorded for this baseline. For tasks with multiple pick--place sub-steps (e.g., stack three cubes), the stages are numbered 1, 2, \ldots in execution order.

\begin{table*}[!htbp]
\caption{\textbf{Sub-task success rate breakdown} (\%) for the 10 multi-stage tasks in the 20-task benchmark. The final stage of each task equals the overall task SR reported in Table~\ref{tab:benchmark}. Best per stage in \textbf{bold}.}
\label{tab:subtask}
\centering
\small
\setlength{\tabcolsep}{4pt}
\definecolor{pickgreen}{rgb}{0.85,0.95,0.85}
\begin{tabular}{ccccccc}
\toprule
\textbf{Task} & \textbf{Stage} & \textbf{BC} & \textbf{DP} & \textbf{ACT} & \textbf{VLA-adapter} & \textbf{Pi\,0.5} \\
\midrule
\multicolumn{7}{c}{\textit{Pick \& Place}} \\
\midrule
\multirow{2}{*}{\textit{pick banana place plate}} & \colorbox{pickgreen}{pick}   & \colorbox{pickgreen}{---} & \colorbox{pickgreen}{90} & \colorbox{pickgreen}{63} & \colorbox{pickgreen}{\textbf{93}} & \colorbox{pickgreen}{\textbf{93}} \\
 & place  & --- & 82 & 47 & 87 & \textbf{93} \\
\midrule
\multirow{2}{*}{\textit{pick cup place shelf}} & \colorbox{pickgreen}{pick}   & \colorbox{pickgreen}{---} & \colorbox{pickgreen}{67} & \colorbox{pickgreen}{87} & \colorbox{pickgreen}{\textbf{100}} & \colorbox{pickgreen}{\textbf{100}} \\
 & place  & --- & 28 & 26 & 55 & \textbf{61} \\
\midrule
\multirow{2}{*}{\textit{pick mug place coffee machine}} & \colorbox{pickgreen}{pick}   & \colorbox{pickgreen}{---} & \colorbox{pickgreen}{63} & \colorbox{pickgreen}{57} & \colorbox{pickgreen}{\textbf{93}} & \colorbox{pickgreen}{\textbf{93}} \\
 & place  & --- & 33 & 27 & 51 & \textbf{53} \\
\midrule
\multirow{2}{*}{\textit{pick lid place pot}} & \colorbox{pickgreen}{pick}   & \colorbox{pickgreen}{---} & \colorbox{pickgreen}{\textbf{53}} & \colorbox{pickgreen}{17} & \colorbox{pickgreen}{\textbf{53}} & \colorbox{pickgreen}{\textbf{53}} \\
 & place  & --- & 37 & 17 & \textbf{53} & 50 \\
\midrule
\multicolumn{7}{c}{\textit{Pour}} \\
\midrule
\multirow{2}{*}{\textit{pour scoop into bowl}} & \colorbox{pickgreen}{pick}   & \colorbox{pickgreen}{---} & \colorbox{pickgreen}{\textbf{87}} & \colorbox{pickgreen}{80} & \colorbox{pickgreen}{70} & \colorbox{pickgreen}{70} \\
 & pour   & --- & 35 & 31 & 62 & \textbf{67} \\
\midrule
\multirow{2}{*}{\textit{pour red mug into bowl}} & \colorbox{pickgreen}{pick}   & \colorbox{pickgreen}{---} & \colorbox{pickgreen}{53} & \colorbox{pickgreen}{50} & \colorbox{pickgreen}{\textbf{93}} & \colorbox{pickgreen}{\textbf{93}} \\
 & pour   & --- & 37 & 30 & 62 & \textbf{68} \\
\midrule
\multirow{2}{*}{\textit{pour kettle into bowl}} & \colorbox{pickgreen}{pick}   & \colorbox{pickgreen}{---} & \colorbox{pickgreen}{73} & \colorbox{pickgreen}{\textbf{80}} & \colorbox{pickgreen}{63} & \colorbox{pickgreen}{63} \\
 & pour   & --- & 33 & 26 & \textbf{56} & 53 \\
\midrule
\multicolumn{7}{c}{\textit{Hang}} \\
\midrule
\multirow{2}{*}{\textit{hang white cup on rack}} & \colorbox{pickgreen}{pick}   & \colorbox{pickgreen}{---} & \colorbox{pickgreen}{87} & \colorbox{pickgreen}{\textbf{100}} & \colorbox{pickgreen}{\textbf{100}} & \colorbox{pickgreen}{\textbf{100}} \\
 & hang   & --- & 23 & 10 & \textbf{47} & 42 \\
\midrule
\multirow{2}{*}{\textit{hang red mug on rack}} & \colorbox{pickgreen}{pick}   & \colorbox{pickgreen}{---} & \colorbox{pickgreen}{77} & \colorbox{pickgreen}{87} & \colorbox{pickgreen}{\textbf{100}} & \colorbox{pickgreen}{\textbf{100}} \\
 & hang   & --- & 23 & 21 & \textbf{36} & 34 \\
\midrule
\multicolumn{7}{c}{\textit{Stack}} \\
\midrule
\multirow{4}{*}{\textit{stack three cubes}} & \colorbox{pickgreen}{pick~1}   & \colorbox{pickgreen}{---} & \colorbox{pickgreen}{73} & \colorbox{pickgreen}{93} & \colorbox{pickgreen}{\textbf{97}} & \colorbox{pickgreen}{93} \\
 & place~1  & --- & 60 & 73 & \textbf{85} & 79 \\
 & \colorbox{pickgreen}{pick~2}   & \colorbox{pickgreen}{---} & \colorbox{pickgreen}{50} & \colorbox{pickgreen}{67} & \colorbox{pickgreen}{\textbf{73}} & \colorbox{pickgreen}{69} \\
 & place~2  & --- & 38 & 36 & \textbf{57} & 47 \\
\bottomrule
\end{tabular}
\end{table*}

The breakdown localizes the bottleneck per category. \textbf{Pick stages} succeed reliably across the benchmark, with VLA-adapter and Pi\,0.5 reaching 90--100\% on most tasks; this confirms that the affordance-guided grasps generated by AffordSim are kinematically and visually executable for modern policy classes.
\textbf{Downstream functional stages} are where most failures concentrate: the pour stage drops to 26--68\%, the hang stage on the rack drops to 10--47\%, and the second placement in stacking drops to 36--57\%. The drop is largest for tasks where the policy must align a fine geometric feature with a target (cup-rim onto the pour opening, mug-handle onto the hook), suggesting that current policy classes can imitate the affordance-grounded contact but struggle to chain it into the precise post-grasp motion.

\section{Sim-to-Real Setup}
\label{app:sim2real}

The real world evaluation uses a Franka FR3 robot arm with a Franka Hand gripper, mounted on a tabletop workspace. Two Intel RealSense D435 cameras provide RGB observations (wrist mounted and third person view at $45^\circ$ elevation). The Depth Anything 3 Gaussian background is reconstructed from 15 photographs of the workspace taken from diverse viewpoints, and full reconstruction settings are listed in Appendix~\ref{app:da3}. All policies are evaluated zero-shot without any real world fine tuning, with 10 trials per task.

\section{Depth Anything 3 Background Reconstruction Details}
\label{app:da3}

This section expands on the cross-viewpoint background substitution introduced in Section~\ref{sec:domain_rand}.
For each real deployment scene we capture 10--20 photographs from diverse viewpoints with a handheld phone camera, and run Depth Anything 3~\citep{lin2025depth} to jointly recover per-view metric depth and camera pose, which are then used to seed a 3D Gaussian Splatting field~\citep{kerbl20233d}.
The optimized field is rendered into the simulator with the intrinsics and resolution of the third-person sensor, producing alternative backgrounds that the visual randomizer composites into the rendered RGB observations.

\paragraph{Reconstruction.}
Photographs are processed by Depth Anything 3 at a working resolution of 504 pixels, and the recovered geometry is anchored to the robot frame by aligning the coordinate origin to the arm base, with a scale factor of $0.8$ and a rigid offset of $[-0.2,\,0.4,\,-0.2]$ in meters.
We run Depth Anything 3 in metric mode so that recovered depth and pose are expressed in true metric units, which removes the need for a separate scale calibration step before handoff to the simulator.
The Depth Anything 3 output initializes a 3D Gaussian Splatting field that we refine for $1000$ optimization iterations, with the structural similarity loss weighted at $0.05$ on top of the photometric loss.
The optimized field is exported as a single GLB asset with debug overlays disabled, and the simulator loads it as a static background of Gaussians during data collection.

\paragraph{Cross-viewpoint rendering.}
At simulation time the Gaussian field is rendered with the third-person RealSense D435 camera intrinsics, $f_x = f_y = 1334.65$, $c_x = 1024$, $c_y = 768$, at a resolution of $1536 \times 2048$, matching the simulator's color sensor.
To produce coverage of the workspace from many directions, we orbit a virtual camera around the canonical pose $[0.8517,\,0.7678,\,0.7521]$ in the robot base frame, sweeping an angular range of $[-120^\circ,\,45^\circ]$ at a step of $15^\circ$.
This yields $12$ novel viewpoints per scene, and each viewpoint is dropped into the simulator as an alternative background render during the visual randomization stage of trajectory collection.

\paragraph{Example scenes.}
Figures~\ref{fig:da3_inputs} and~\ref{fig:da3_recon} together illustrate the pipeline on three deployment scenes, namely a coffee corner, a factory bench, and a library desk.
Figure~\ref{fig:da3_inputs} shows five representative phone photographs per scene, drawn from the set of 14--20 captures that is fed to Depth Anything 3.
Figure~\ref{fig:da3_recon} then shows the resulting 3D Gaussian field rendered at the canonical viewpoint together with four orbit angles spanning the supported range.
The same Depth Anything 3 hyperparameters and orbit schedule are used for all three scenes, and the only per-scene input is the photograph set captured at deployment time.

\figDAThreeInputs

\figDAThreeRecon

\section{Scene Graph Generation Prompt}
\label{app:scene_graph_prompt}

The VLM-based scene generator (Section~\ref{sec:vlm_generation}) is driven by the prompt below, which converts a natural-language scene description into a structured JSON scene graph.

\begin{lstlisting}[style=promptstyle]
You are a Scene Graph Parser.

Convert the user's scene description into a JSON object with exactly this schema:
{
  "specified_objects": ["..."],
  "random_objects_count": 0,
  "relationships": [
    {"child": "...", "relation": "...", "parent": "..."}
  ]
}

Rules:
1. Output JSON only. No markdown. No explanation.
2. `specified_objects` must contain only short object reference phrases. Do not include verbs, conjunctions,
   politeness words, or full clauses. Good: "banana", "yellow bowl", "microwave". Bad: "the kettle and then pour".
3. `child` and `parent` must also be short object reference phrases only.
4. Do not invent extra disambiguating attributes that the user did not say. Never add colors, materials, sizes,
   indices, or serial numbers on your own. If the user says "kettle", output "kettle", not "blue_kettle".
5. If the user mentions a count of concrete named objects, repeat that object phrase in `specified_objects`
   exactly that many times and use the singular noun. Example: "two kettles" -> ["kettle", "kettle"].
   Do not put the extra counted concrete objects in `random_objects_count`.
6. Put a number in `random_objects_count` only when the user explicitly asks for random/any/arbitrary objects,
   such as "two random objects". Otherwise use 0.
7. Do not duplicate an uncounted object mention multiple times in `specified_objects`; counted concrete objects
   are the only exception.
8. Output relation values using only these canonical relations:
   - "inside", "on_top"
   - "left_of", "right_of", "front", "behind"
   - "near"
9. Normalize common phrasing to the canonical relations above:
   - in / into / inside -> "inside"
   - on / onto / on top of / atop -> "on_top"
   - left / on the left of / to the left of -> "left_of"
   - right / on the right of / to the right of -> "right_of"
   - in front of / ahead of -> "front"
   - back of / behind / in back of -> "behind"
   - near / next to / close to / beside -> "near"
10. Ignore filler words like "please", "carefully", "then", "after that", "could you".
11. If a relation is not clearly expressed, omit it instead of inventing one.

Example input:
"Please create a plate with a banana inside, and a fork on the left of the plate."

Example output:
{
  "specified_objects": ["plate", "banana", "fork"],
  "random_objects_count": 0,
  "relationships": [
    {"child": "banana", "relation": "inside", "parent": "plate"},
    {"child": "fork", "relation": "left_of", "parent": "plate"}
  ]
}

Example input:
"two kettles"

Example output:
{
  "specified_objects": ["kettle", "kettle"],
  "random_objects_count": 0,
  "relationships": []
}

Example input:
"one kettle and two random objects"

Example output:
{
  "specified_objects": ["kettle"],
  "random_objects_count": 2,
  "relationships": []
}
\end{lstlisting}

\section{Affordance Query Extraction Prompt}
\label{app:affordance_query_prompt}

The VLM-powered task generation step (Section~\ref{sec:vlm_generation}) parses each natural-language task into a sequence of action primitives and per-object affordance queries using the prompt below. Worked examples cover pick--pour--place chains, multi-object pick sequences, absolute-pose placement, and articulated-object opening.

\begin{lstlisting}[style=promptstyle]
You are a task primitive parser.

Convert the user's natural-language task into a JSON object with this format:
{
  "primitives": [
    {"type": "...", "target": "...", "mode": "...", "position_xy": [x, y],
     "affordance_queries": [{"object": "...", "action": "..."}]}
  ]
}

Rules:
1. Allowed primitive types:
   pick, place, pour, push, pull, open, close, hang, hold, return_home
2. Output JSON only. No markdown. No explanation.
3. Each `target` must be only a short object reference phrase. Never include conjunctions, transition words,
   or the next action in `target`. Good: "kettle", "bowl", "plate". Bad: "kettle and then pour in the bowl".
4. Never invent extra disambiguating attributes that the user did not say. Do not add colors, materials, sizes,
   indices, or serial numbers on your own. If the user says "kettle", output "kettle", not "blue_kettle" or
   "white_kettle10". Only use an exact label when the user's wording already uniquely specifies it, such as
   "blue kettle" or "white kettle 10".
5. `target` is required for: pick, place(in/on), pour, push, pull, open, close, hang.
6. `target` must be omitted for: place(pose), hold, return_home.
7. Only `pick` and `place` may include `mode`.
8. If pick mode is not explicitly specified, use "affordance".
9. If place mode says in/into/inside, use "in".
10. If place mode says on/onto/on top of, use "on".
11. If the user specifies an absolute placement location such as "place at x=..., y=..." or "place to (x, y)",
    use {"type": "place", "mode": "pose", "position_xy": [x, y]} and omit `target`.
12. If place relation is not explicitly specified, default to "on".
13. For push, output `position_xy: [x, y]` only when the instruction explicitly provides x and y.
14. For place pose, output `position_xy: [x, y]` only when the instruction explicitly provides x and y.
15. Do not output `position_xy` for non-push, non-place-pose primitives.
16. Preserve execution order exactly as implied by the instruction.
17. Ignore filler words like "please", "then", "after that", "carefully", "finally".

Affordance query rules:
A1. `affordance_queries` is a list of {"object": "<short label>", "action": "<verb>"} entries describing
    which objects need an affordance region annotated and for which action.
A2. The only allowed `action` values are:
    "pick", "place", "pour", "push", "pull", "open", "close", "hang".
    Never use any other verb (no "grasp", "receive", "lift", etc.).
A3. The `object` field must be a short object reference phrase, following the same rules as `target`.
    Never invent disambiguating attributes the user did not say.
A4. Emit `affordance_queries` for: pick, place(in/on), pour, push, pull, open, close, hang.
    Do NOT emit `affordance_queries` for: place(pose), hold, return_home (omit the field entirely).
A5. For pick, push, pull, open, close, hang, place(in/on): output exactly one entry whose `object`
    equals the primitive's `target`, and whose `action` equals the primitive's `type`.
A6. For pour: output exactly two entries, both with `action: "pour"`:
    - one with `object` = the in-hand object (the most recent prior `pick`'s target),
    - one with `object` = the pour primitive's `target` (the receiving container).
    If no prior pick is visible in the instruction, emit only the receiving-container entry.
A7. Preserve the order of entries as: in-hand object first, then target container (for pour); target only
    (for the other action types).

Example input:
"Pick the kettle and then pour in the bowl, then place the kettle on the plate."

Example output:
{
  "primitives": [
    {"type": "pick", "target": "kettle", "mode": "affordance",
     "affordance_queries": [{"object": "kettle", "action": "pick"}]},
    {"type": "pour", "target": "bowl",
     "affordance_queries": [
       {"object": "kettle", "action": "pour"},
       {"object": "bowl", "action": "pour"}
     ]},
    {"type": "place", "target": "plate", "mode": "on",
     "affordance_queries": [{"object": "plate", "action": "place"}]}
  ]
}

Example input:
"Pick the kettle, then pick the cup."

Example output:
{
  "primitives": [
    {"type": "pick", "target": "kettle", "mode": "affordance",
     "affordance_queries": [{"object": "kettle", "action": "pick"}]},
    {"type": "pick", "target": "cup", "mode": "affordance",
     "affordance_queries": [{"object": "cup", "action": "pick"}]}
  ]
}

Example input:
"Pick the mug and then place it at x = 0.5, y = -0.4."

Example output:
{
  "primitives": [
    {"type": "pick", "target": "mug", "mode": "affordance",
     "affordance_queries": [{"object": "mug", "action": "pick"}]},
    {"type": "place", "mode": "pose", "position_xy": [0.5, -0.4]}
  ]
}

Example input:
"Open the drawer."

Example output:
{
  "primitives": [
    {"type": "open", "target": "drawer",
     "affordance_queries": [{"object": "drawer", "action": "open"}]}
  ]
}
\end{lstlisting}

% ==================== CHECKLIST ====================

\newpage
\section*{NeurIPS Paper Checklist}

\begin{enumerate}

\item {\bf Claims}
    \item[] Question: Do the main claims made in the abstract and introduction accurately reflect the paper's contributions and scope?
    \item[] Answer: \answerYes{}
    \item[] Justification: The abstract and Section~\ref{sec:introduction} state three contributions (C1--C3): (1) the AffordSim framework, (2) the 50-task affordance-aware benchmark, and (3) zero-shot sim-to-real transfer of VLA policies trained on AffordSim data. Each is supported by experiments in Section~\ref{sec:experiments} (Tables~\ref{tab:afford_main}, \ref{tab:benchmark}, and \ref{tab:sim2real}).

\item {\bf Limitations}
    \item[] Question: Does the paper discuss the limitations of the work performed by the authors?
    \item[] Answer: \answerYes{}
    \item[] Justification: Appendix~\ref{app:limitations} discusses three limitations: (i) VoxAfford predictor coverage on novel geometries, transparent objects, and ambiguous functional regions; (ii) restriction to rigid bodies, with deformables, in-hand, and bimanual manipulation deferred to future work; (iii) Depth Anything 3 background reconstruction requires 10--20 photographs per scene and degrades sharply at viewpoints far from the captured poses.

\item {\bf Theory assumptions and proofs}
    \item[] Question: For each theoretical result, does the paper provide the full set of assumptions and a complete (and correct) proof?
    \item[] Answer: \answerNA{}
    \item[] Justification: The paper does not include theoretical results; it is a systems and benchmark contribution.

    \item {\bf Experimental result reproducibility}
    \item[] Question: Does the paper fully disclose all the information needed to reproduce the main experimental results of the paper to the extent that it affects the main claims and/or conclusions of the paper (regardless of whether the code and data are provided or not)?
    \item[] Answer: \answerYes{}
    \item[] Justification: Section~\ref{sec:experiments} describes the experimental setup, baselines, and evaluation protocol. Appendix~\ref{app:training} provides training hyperparameters for all baselines, Appendix~\ref{app:manual_baseline} specifies the Manual baseline rules, and Appendix~\ref{app:sim2real} together with Appendix~\ref{app:da3} detail the sim-to-real hardware setup and Depth Anything 3 reconstruction.

\item {\bf Open access to data and code}
    \item[] Question: Does the paper provide open access to the data and code, with sufficient instructions to faithfully reproduce the main experimental results, as described in supplemental material?
    \item[] Answer: \answerYes{}
    \item[] Justification: We will release the AffordSim codebase, benchmark tasks, and generated datasets under a permissive open-source license upon publication. An anonymized version is provided during review.

\item {\bf Experimental setting/details}
    \item[] Question: Does the paper specify all the training and test details (e.g., data splits, hyperparameters, how they were chosen, type of optimizer) necessary to understand the results?
    \item[] Answer: \answerYes{}
    \item[] Justification: Section~\ref{sec:experiments} and Appendix~\ref{app:training} specify all training details including architectures, hyperparameters, optimizers, learning rates, batch sizes, and training epochs for each baseline (BC, Diffusion Policy, ACT, VLA-adapter, Pi\,0.5).

\item {\bf Experiment statistical significance}
    \item[] Question: Does the paper report error bars suitably and correctly defined or other appropriate information about the statistical significance of the experiments?
    \item[] Answer: \answerYes{}
    \item[] Justification: Each task is evaluated over 100 simulation rollouts for trajectory collection (100 for benchmark policy evaluation, 10 for real-robot experiments). Success rates are reported as percentages. We report per-task results in addition to aggregated tier and category averages, allowing readers to assess variance across tasks.

\item {\bf Experiments compute resources}
    \item[] Question: For each experiment, does the paper provide sufficient information on the computer resources (type of compute workers, memory, time of execution) needed to reproduce the experiments?
    \item[] Answer: \answerYes{}
    \item[] Justification: Appendix~\ref{app:training} specifies that all imitation-learning experiments use a single node with 8 NVIDIA A800 80GB GPUs and reports per-task training times (2--26 hours depending on the baseline).

\item {\bf Code of ethics}
    \item[] Question: Does the research conducted in the paper conform, in every respect, with the NeurIPS Code of Ethics \url{https://neurips.cc/public/EthicsGuidelines}?
    \item[] Answer: \answerYes{}
    \item[] Justification: The research involves simulation-based and tabletop robotic manipulation with no human subjects, no sensitive data, and no IRB-relevant procedures.

\item {\bf Broader impacts}
    \item[] Question: Does the paper discuss both potential positive societal impacts and negative societal impacts of the work performed?
    \item[] Answer: \answerYes{}
    \item[] Justification: Beyond data collection, AffordSim also serves as a systematic evaluation suite: by curating affordance-critical tasks, it measures how well a policy actually understands object affordance rather than merely succeeds on grasp-stable scenarios, an evaluation that has been out of reach for existing platforms which lack benchmarks for affordance-dependent tasks. Positive impacts therefore include accelerating affordance-aware manipulation research, exposing failure modes that grasp-stability-only benchmarks hide, and enabling more reliable deployment in manufacturing, logistics, and assistive robotics; policies trained with AffordSim data further exhibit zero-shot sim-to-real transfer, demonstrating the practical utility of the collected trajectories. Potential negative impacts include enabling autonomous manipulation systems that may be deployed without sufficient human oversight or that displace manual labor; we view these as inherent to robotic-manipulation research rather than uniquely raised by this work, and the framework imposes no application-specific deployment.

\item {\bf Safeguards}
    \item[] Question: Does the paper describe safeguards that have been put in place for responsible release of data or models that have a high risk for misuse (e.g., pre-trained language models, image generators, or scraped datasets)?
    \item[] Answer: \answerNA{}
    \item[] Justification: The released assets (simulation tasks, benchmark data, evaluation code) do not pose risks for misuse beyond standard robotic manipulation research.

\item {\bf Licenses for existing assets}
    \item[] Question: Are the creators or original owners of assets (e.g., code, data, models), used in the paper, properly credited and are the license and terms of use explicitly mentioned and properly respected?
    \item[] Answer: \answerYes{}
    \item[] Justification: All external assets (VoxAfford, AnyGrasp, Isaac Sim, Depth Anything 3, Qwen-VL-Plus, baseline policies BC/DP/ACT/VLA-adapter/Pi\,0.5) are cited in the main paper and bibliography. License information will be provided in the code release.

\item {\bf New assets}
    \item[] Question: Are new assets introduced in the paper well documented and is the documentation provided alongside the assets?
    \item[] Answer: \answerYes{}
    \item[] Justification: The AffordSim framework, benchmark tasks (Appendix~\ref{app:benchmark_categories}, Appendix~\ref{app:tier_listings}), and generated datasets are released at \url{https://anonymous.4open.science/r/AffordSim-8D94/} with documentation including installation instructions, task descriptions, and evaluation scripts.

\item {\bf Crowdsourcing and research with human subjects}
    \item[] Question: For crowdsourcing experiments and research with human subjects, does the paper include the full text of instructions given to participants and screenshots, if applicable, as well as details about compensation (if any)?
    \item[] Answer: \answerNA{}
    \item[] Justification: The paper does not involve crowdsourcing or research with human subjects.

\item {\bf Institutional review board (IRB) approvals or equivalent for research with human subjects}
    \item[] Question: Does the paper describe potential risks incurred by study participants, whether such risks were disclosed to the subjects, and whether Institutional Review Board (IRB) approvals (or an equivalent approval/review based on the requirements of your country or institution) were obtained?
    \item[] Answer: \answerNA{}
    \item[] Justification: The paper does not involve research with human subjects.

\item {\bf Declaration of LLM usage}
    \item[] Question: Does the paper describe the usage of LLMs if it is an important, original, or non-standard component of the core methods in this research? Note that if the LLM is used only for writing, editing, or formatting purposes and does \emph{not} impact the core methodology, scientific rigor, or originality of the research, declaration is not required.
    \item[] Answer: \answerYes{}
    \item[] Justification: Section~\ref{sec:vlm_generation} describes the use of Qwen-VL-Plus as the VLM backbone for separate scene and task parsing followed by affordance-query conversion, which is a core component of the AffordSim pipeline. The full prompts are provided in Appendix~\ref{app:scene_graph_prompt} and Appendix~\ref{app:affordance_query_prompt}.

\end{enumerate}

\end{document}